\documentclass{article}

     \PassOptionsToPackage{numbers, compress}{natbib}
 \usepackage[preprint]{neurips_2026}

\usepackage[utf8]{inputenc} 
\usepackage[T1]{fontenc}    
\usepackage{hyperref}       
\usepackage{url}            
\usepackage{booktabs}       
\usepackage{amsfonts}       
\usepackage{nicefrac}       
\usepackage{microtype}      
\usepackage{xcolor}         
\usepackage{graphicx}       
\usepackage{amsmath}
\usepackage{wrapfig}
\usepackage{multirow}
\usepackage{xcolor}

\usepackage{bm}
\usepackage[table]{xcolor} 
\usepackage{caption}
\definecolor{bestcolor}{HTML}{FFC1C1}   
\definecolor{secondcolor}{HTML}{FFE4C4} 
\definecolor{thirdcolor}{HTML}{FFFFCC}  

\newcommand{\first}[1]{\cellcolor{bestcolor}\textbf{#1}}
\newcommand{\second}[1]{\cellcolor{secondcolor}#1}
\newcommand{\third}[1]{\cellcolor{thirdcolor}#1}
\title{WebSpline: Structure-Informed Splines \\ for Real-Time 3D Gaussians from Monocular Videos}

\author{Jongmin Park\footnotemark[1] \quad Jeonghwan Yun\footnotemark[1] \quad Minh-Quan Viet Bui \quad Munchurl Kim\\
KAIST\\
\texttt{\{jm.park, jeonghwan.yun, bvmquan, mkimee\}@kaist.ac.kr} \\
\url{https://kaist-viclab.github.io/webspline-site/}
\vspace{-0.4cm}
}

\makeatletter
\renewcommand{\@noticestring}{}
\makeatother

\begin{document}

\maketitle

\begin{figure}[h]
    \centering
    \includegraphics[width=\linewidth,keepaspectratio]{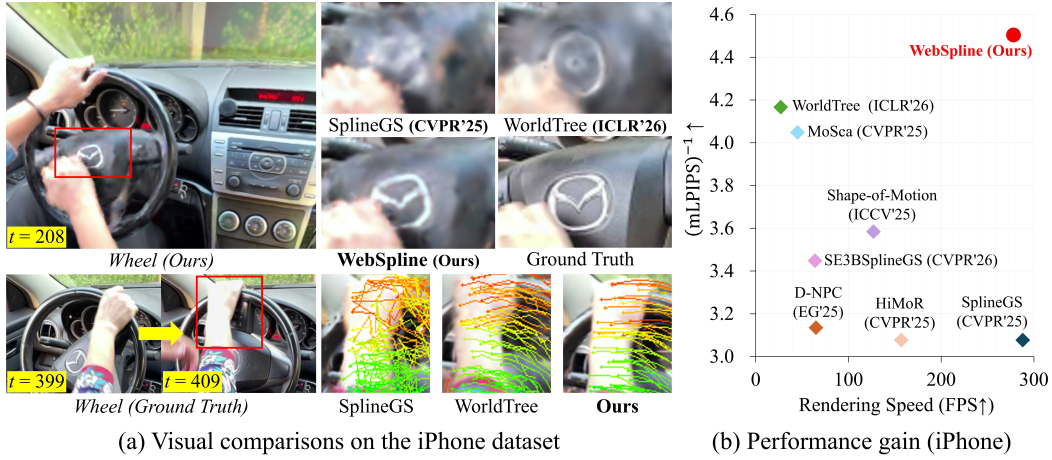}
    \vspace{-0.3cm}
    \caption{\textbf{WebSpline achieves high-quality dynamic scene reconstruction with fast rendering from monocular videos.} (a) Qualitative comparison with state-of-the-art methods on novel view synthesis (top) and visualized Gaussian trajectories (bottom). (b) WebSpline achieves the best rendering quality while rendering over 10$\times$ faster than the second-best method.}
    \label{fig:figure_page1}
\end{figure}

{
  \renewcommand{\thefootnote}%
    {\fnsymbol{footnote}}
  \footnotetext[1]{Co-first authors (equal contribution).}
}

\begin{abstract}
Dynamic scene reconstruction from monocular videos remains highly challenging, as existing methods often struggle to balance global structural coherence and local fine-grained details under limited multi-view cues. To address this challenge, we propose WebSpline, a novel dynamic 3D Gaussian framework that enables structurally coherent and high-fidelity reconstruction from monocular videos with fast rendering. The core of WebSpline is the Structure-Informed Spline (SIS) representation, which models each dynamic Gaussian trajectory using a learnable cubic Hermite spline whose motion is structurally organized with an auxiliary Structural Proxy Graph (SPG). The proposed framework is optimized in two stages: (i) in the first stage, the SPG is initialized from 2D point tracks and refined with temporal rigidity regularization to establish structural coherence for moving objects across the sequence; and (ii) in the second stage, the SIS representation is initialized from the refined SPG and optimized under both spatial and structural neighborhood constraints. At inference, Gaussian motion is obtained solely by evaluating the learned SIS, enabling fast rendering. Extensive experiments on the challenging monocular dynamic scene benchmarks, iPhone and NVIDIA, demonstrate that our WebSpline achieves \textbf{state-of-the-art rendering quality} while rendering over \textbf{10$\times$ faster} than WorldTree, the second-best method on the iPhone dataset.
\end{abstract}

\section{Introduction}
\label{sec:intro}

Dynamic scene reconstruction from monocular videos is a fundamental task in 3D vision, with broad applications in virtual reality (VR), robotics, and autonomous driving. High-fidelity reconstruction in this setting requires modeling complex object deformations and time-varying geometry from limited observations while preserving spatio-temporal consistency. Recently, 3D Gaussian Splatting (3DGS)~\cite{3dgs} has established a powerful paradigm for scene reconstruction by representing a scene with learnable 3D Gaussians and rendering them via differentiable splatting, which enables high-quality, real-time rendering. However, since 3DGS~\cite{3dgs, keselman2022approximate, keselman2023flexible, yu2024gaussian, yu2024mip} is primarily designed for static scenes, extending it to monocular dynamic scene reconstruction remains highly challenging.

To tackle these challenges, several recent methods~\cite{4dgs, huang2023sc, deformable3dgs, mosca, worldtree, splinegs, som, bui2026mobgs, qingming2025modgs, lin2024gaussian} have extended 3DGS~\cite{3dgs} to dynamic scene reconstruction for monocular videos by modeling the time-varying deformations of 3D Gaussians. Recent state-of-the-art (SOTA) methods can be categorized by their Gaussian motion modeling strategies: (i) scaffold-based methods~\cite{mosca, worldtree, se3bsplinegs} and (ii) primitive-motion-parameterized methods~\cite{splinegs, som}. The scaffold-based methods~\cite{mosca, worldtree, se3bsplinegs} model Gaussian motions through scaffold structures, where the trajectory of each Gaussian is determined by blending the motions of neighboring scaffolds over time. While effective in preserving structural consistency, such blending often introduces over-smoothing artifacts, resulting in blurry renderings and degraded perceptual quality. In contrast, the primitive-motion-parameterized methods~\cite{splinegs, som} model per-Gaussian motion using learnable trajectory parameters such as spline functions~\cite{splinegs} or $SE(3)$ motion coefficients~\cite{som}. These methods can produce sharper renderings, but the motion of each Gaussian is less constrained, often leading to geometric instability for moving objects.

To address these limitations, we propose WebSpline, a novel dynamic Gaussian framework for structurally coherent and high-fidelity monocular dynamic scene reconstruction with fast rendering. At the core of WebSpline is a structurally organized spline representation that models each dynamic Gaussian with flexible per-primitive trajectories while leveraging an auxiliary graph structure during training. WebSpline consists of two components: (i) a Structural Proxy Graph (SPG) and (ii) a Structure-Informed Spline (SIS) representation, optimized through a two-stage framework. In the first stage, we construct the SPG initialized from 2D point tracks and optimize it to serve as a robust structural foundation. To achieve this, we introduce an orientation-similarity-based temporal rigidity loss to refine the SPG node trajectories for structural coherence across the sequence. In the second stage, guided by the optimized SPG, the SIS is introduced for parameterized representation of each dynamic Gaussian trajectory using a cubic Hermite spline~\cite{ahlberg2016theory, de1978practical} with learnable control points. Within SIS, each dynamic Gaussian is associated with a corresponding SPG node, aligning its spline trajectory with the structurally organized SPG node trajectory. During SIS optimization, we define two types of neighborhoods, spatial and structural, and impose temporal rigidity constraints over these neighborhoods.
These rigidity constraints stabilize SIS learning by preserving both local geometric consistency and structural coherence over time, while retaining the flexibility needed to capture fine-grained non-rigid motion.
At inference, the SPG is no longer required, and Gaussian motion is obtained solely by evaluating the learned SIS, enabling fast rendering. Notably, while SplineGS~\cite{splinegs} also represents dynamic Gaussian motion with spline trajectories, WebSpline incorporates spatial and structural neighborhood constraints, thereby reducing noisy artifacts and promoting more reliable geometry. Extensive experiments show that WebSpline significantly outperforms the SOTA dynamic Gaussian methods, as shown in Table~\ref{tab:comparison_results} and Figs.~\ref{fig:figure_page1}, \ref{fig:qualitative_iphone}, and \ref{fig:qualitative_nvidia}. WebSpline achieves PSNR gains of \textbf{+0.36 dB} and \textbf{+0.61 dB} over the second-best methods on the iPhone~\cite{gao2022monocular} and NVIDIA~\cite{yoon2020dynamic} datasets, respectively. Furthermore, WebSpline enables \textbf{10$\times$ faster rendering} than WorldTree~\cite{worldtree}, the second-best method on the iPhone dataset. Our contributions are as follows:
\begin{itemize}
\item We propose \textbf{WebSpline}, a novel approach for dynamic 3D Gaussians that enables structurally coherent and high-fidelity reconstruction for monocular videos with fast rendering.

\item We introduce the \textbf{Structure-Informed Spline (SIS)}, which parameterizes fine-grained Gaussian trajectories under the structural organization of a \textbf{Structural Proxy Graph (SPG)} for structurally coherent dynamic motion modeling.

\item We present a two-stage optimization strategy that first constructs the SPG as a structural foundation and then learns the SIS representation for structurally coherent spline motion.

\item Our WebSpline achieves SOTA performance in all metrics on challenging monocular benchmarks, while rendering over $\mathbf{10\times}$ faster than the second-best method on the iPhone dataset.
\end{itemize}  
\section{Related Work}
\label{sec:related_work}

\textbf{3D Scene Reconstruction.} The emergence of neural radiance fields (NeRF)~\cite{nerf, mipnerf, mipnerf360} has significantly advanced 3D scene reconstruction and novel view synthesis. NeRF~\cite{nerf, mipnerf, mipnerf360} represents continuous 3D scenes using an MLP-based implicit representation and renders them via volumetric rendering. This process requires dense ray sampling and repeated MLP evaluations along each ray, resulting in substantial computation overhead during both training and inference. Although numerous NeRF-based approaches~\cite{chen2022tensorf, sun2022direct, wang2022fourier, fridovich2022plenoxels, muller2022instant, yu2021plenoctrees} have been proposed to accelerate rendering, they often struggle to achieve a favorable balance between rendering fidelity and efficiency. Recently, 3D Gaussian Splatting (3DGS)~\cite{3dgs} has emerged as an explicit scene representation based on anisotropic 3D Gaussians, together with a differentiable rasterization pipeline. This design enables real-time rendering while maintaining high-quality scene reconstruction. However, 3DGS~\cite{3dgs} is primarily designed for static scene reconstruction and lacks a temporal modeling capability, which limits its applicability for dynamic scene reconstruction. 

\paragraph{Dynamic 3D Gaussian Splatting.}
To extend 3DGS to dynamic scenes, prior works have explored diverse strategies, including canonical-space methods~\cite{deformable3dgs, liang2023gaufre, wan2024superpoint, 4dgs, st4dgs, bae2024per}, explicit-state methods~\cite{dyn3dgs, Li_STG_2024_CVPR, lee2024fully}, and structured motion learning methods~\cite{huang2023sc, modecgs, liang2025himor, stearns2024dynamic}. More recently, SOTA methods~\cite{mosca, worldtree, se3bsplinegs, som, splinegs} tailored for monocular video settings have been proposed. These methods incorporate various priors, such as optical flows~\cite{teed2020raft, xu2022gmflow}, monocular depths~\cite{piccinelli2024unidepth, depthanything, lin2025depth, yang2024depth}, motion masks~\cite{sam}, or 2D point tracks~\cite{karaev2023cotracker, karaev2025cotracker3}, to compensate for the limited multi-view cues and improve geometric accuracy. These methods can be divided into two groups: (i) scaffold-based methods~\cite{mosca, worldtree, se3bsplinegs} and (ii) primitive-motion-parameterized methods~\cite{splinegs, som}. For the scaffold-based methods, MoSca~\cite{mosca} optimizes a scaffold architecture from 2D point tracks and represents Gaussian trajectories by blending the motions of neighboring nodes within the scaffold. WorldTree~\cite{worldtree} builds upon the scaffold-based framework of MoSca~\cite{mosca} by adopting a hierarchical tree-chain structure. SE3BSplineGS~\cite{se3bsplinegs} extends MoSca~\cite{mosca} by adopting B-spline basis functions~\cite{de1978practical}. However, as these methods rely on motion blending across scaffold nodes, the rendering results tend to be blurry, and the inference speed is relatively slow. For the primitive-motion-parameterized methods, SoM~\cite{som} decomposes Gaussian motion into a set of shared $SE(3)$ motion bases and optimizes coefficients for each Gaussian primitive. SplineGS~\cite{splinegs} parameterizes each Gaussian trajectory using a cubic Hermite spline function~\cite{ahlberg2016theory, de1978practical} to ensure smooth and continuous motion while enabling high rendering speed. However, as these approaches lack explicit spatial constraints between primitives, they often struggle to maintain structural coherence and may produce noisy artifacts.
Unlike these methods, our proposed WebSpline integrates a Structural Proxy Graph (SPG) as a robust structural motion foundation with a Structure-Informed Spline (SIS) representation for flexible per-primitive trajectory modeling. This design achieves structural coherence while preserving rich motion details and enabling fast rendering.
\section{Method}
\label{sec:method}
\subsection{Overview of WebSpline}
\begin{figure*}
\centering
\includegraphics[width=\linewidth,keepaspectratio]{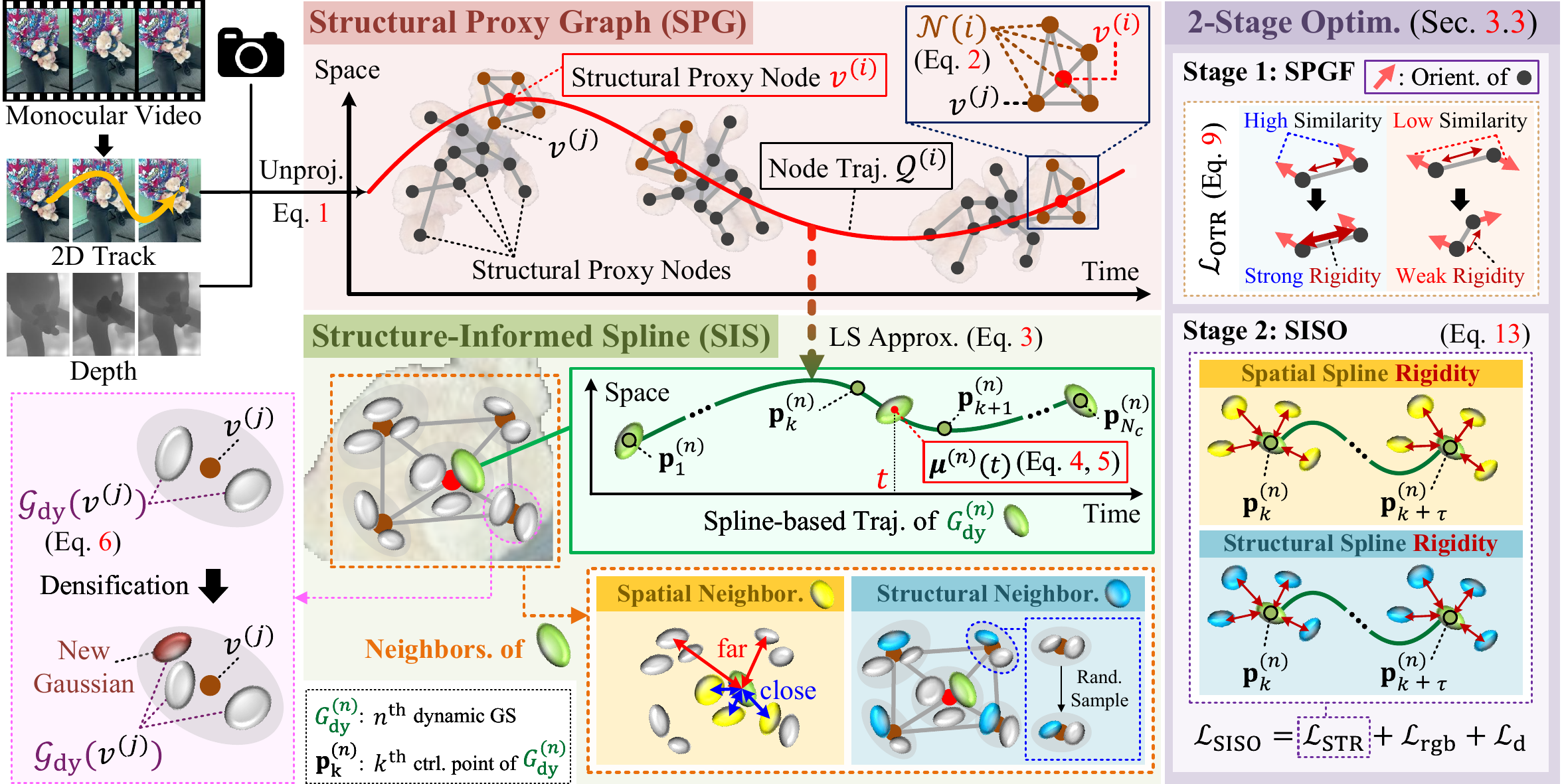}
\vspace{-0.4cm}
\caption{\textbf{Overview of WebSpline.} 
WebSpline models each dynamic Gaussian trajectory using the Structure-Informed Spline (SIS) representation, initialized from the Structural Proxy Graph (SPG). 
For SIS optimization, we define two types of neighborhoods for each dynamic Gaussian, spatial and structural, to enforce coherent spline motion while capturing fine-grained dynamics.}
\label{fig:overall_architecture}
\vspace{-0.5cm}
\end{figure*}

Fig.~\ref{fig:overall_architecture} illustrates the overall architecture of WebSpline. Given a monocular video $\{\bm{I}_t\}_{t=1}^{N_f}$ of $N_f$ frames, WebSpline aims to achieve high-quality scene reconstruction by jointly capturing structurally coherent and fine-grained motions of dynamic objects, while enabling fast inference. We represent the scene using a set $\mathcal{G}$ of 3D Gaussians, which is decomposed into static Gaussians $\mathcal{G}_{\text{st}}$ and dynamic Gaussians $\mathcal{G}_{\text{dy}}$, where $N_g = |\mathcal{G}_{\text{dy}}|$ denotes the number of dynamic Gaussians. For each dynamic Gaussian $G_{\text{dy}}^{(n)} \in \mathcal{G}_{\text{dy}}$, we model its motion using a structurally organized spline-based representation, optimized through a two-stage framework: (i) in the Structural Proxy Graph Foundation (SPGF) stage, we construct a Structural Proxy Graph (SPG) $\mathcal{K}$ from 2D point tracks~\cite{karaev2025cotracker3} to provide structural guidance; and (ii) in the Structure-Informed Spline Optimization (SISO) stage, building on the SPG $\mathcal{K}$, we parameterize the motion of each dynamic Gaussian using a Structure-Informed Spline (SIS) representation, where each dynamic Gaussian is assigned to a node in the SPG and its trajectory is modeled by a learnable cubic Hermite spline~\cite{ahlberg2016theory, de1978practical}. During SIS optimization, we impose temporal rigidity constraints over two complementary neighborhoods: spatial neighborhoods defined by nearby Gaussian trajectories and structural neighborhoods induced by the SPG connectivity. Notably, the SPG $\mathcal{K}$ is used only during training to enforce structural coherence, while dynamic Gaussian motion during inference is obtained solely by evaluating the learned SIS, enabling fast rendering.

\subsection{Structure-Informed Splines (SIS) for Dynamic 3D Gaussians}
\textbf{Structural Proxy Graph (SPG).}
To improve structural coherence in spline-based dynamic Gaussian representation of our WebSpline, we introduce the SPG $\mathcal{K} = (\mathcal{V}, \mathcal{E})$, where $\mathcal{V}$ denotes the set of $M$ structural proxy nodes and $\mathcal{E}$ denotes the edge set connecting neighboring nodes. As depicted in Fig.~\ref{fig:overall_architecture}, the SPG provides a structural motion representation of dynamic components in a scene, with each structural proxy node $v^{(i)} \in \mathcal{V}$ representing a local dynamic region through its trajectory. Each $v^{(i)}$ is associated with a learnable structural proxy node trajectory $\mathcal{Q}^{(i)} = \{\mathbf{q}_t^{(i)}\}_{t=1}^{N_f}$ and a learnable auxiliary orientation sequence $\mathcal{O}^{(i)} = \{\mathbf{O}_t^{(i)}\}_{t=1}^{N_f}$, where $\mathbf{q}_t^{(i)} \in \mathbb{R}^3$ denotes its position and $\mathbf{O}_t^{(i)} \in SO(3)$ denotes its auxiliary orientation at time $t$.
To initialize $\mathcal{Q}^{(i)}$, we first lift 2D point tracks extracted from dynamic regions of the scene into 3D using camera extrinsics and metric depth.
Let $\mathcal{T}^{(i)} = \{\mathbf{u}_t^{(i)}\}_{t=1}^{N_f}$ denote the $i^\text{th}$ 2D point track, where $\mathbf{u}_t^{(i)}$ is the tracked pixel coordinate at time $t$. Given the metric depth $\bm d_t \in \mathbb{R}^{H \times W}$ and camera extrinsics 
$[\bm R_t |\bm T_t] \in \mathbb{R}^{3 \times 4}$ at time $t$, we initialize the 3D position of each structural proxy node by lifting the tracked pixel $\mathbf{u}_t^{(i)}$ as
\begin{equation}
\mathbf{q}_t^{(i)} = \bm R_t^\top 
\pi_{\bm K}^{-1}(\mathbf{u}_t^{(i)}, \bm d_t(\mathbf{u}_t^{(i)}))
- \bm R_t^\top \bm T_t,
\label{eq:unproj}
\end{equation}
where $\pi_{\bm K}^{-1}$ denotes the inverse projection from image to camera coordinates under intrinsics $\bm K$. The initialization of $\mathcal{O}^{(i)}$ is described in the Appendix~\ref{subsec:orient_init}. We then determine $\mathcal{E}$ using a pairwise trajectory distance matrix 
$\mathcal{D} \in \mathbb{R}^{M \times M}$, where each entry is defined as 
$\mathcal{D}_{ij}=Q_{1-\eta}(\{\|\mathbf{q}_t^{(i)}-\mathbf{q}_t^{(j)}\|_2\}_{t=1}^{N_f})$.
Here, $Q_{1-\eta}$ denotes the $(1-\eta)$-quantile over time. The neighborhood $\mathcal{N}(i)$ of each structural proxy node $v^{(i)}$ and the edge set $\mathcal{E}$ are defined as
\begin{equation}
\mathcal{N}(i) = \{ 
j \in \{1,\dots, M\} 
\;|\; 
j \neq i, \; \operatorname{rank}_i(\mathcal{D}_{ij}) \le K 
\},
\quad
\mathcal{E} = \bigcup_{i=1}^{M}\{(v^{(i)}, v^{(j)}) \mid j \in \mathcal{N}(i)\},
\label{eq:neighbor_and_edges}
\end{equation}
where $K$ denotes the number of neighboring nodes of $v^{(i)}$ and $\operatorname{rank}_i(\mathcal{D}_{ij})$ denotes the ascending rank of $\mathcal{D}_{ij}$ among all distances from $v^{(i)}$ to others. 
To ensure that the initialized SPG $\mathcal{K}$ provides robust motion guidance, it is further optimized for structural coherence over time, as detailed in Sec.~\ref{subsec:optimization}.

\textbf{SIS for Gaussian Motion.}
Building on the SPG $\mathcal{K}$, we propose the SIS representation for the motion of each dynamic 3D Gaussian, combining structural motion coherence with fine-grained deformation modeling. The motion of $G_{\text{dy}}^{(n)}$ is parameterized by the pair $(c^{(n)}, \mathbf{P}^{(n)})$, where $c^{(n)} \in \{1,\dots,M\}$ denotes the node index of the associated structural proxy node and $\mathbf{P}^{(n)} = \{\mathbf{p}_k^{(n)}\}_{k=1}^{N_c}$ denotes a learnable control point set of a cubic Hermite spline representing the trajectory of $G_{\text{dy}}^{(n)}$. Specifically, $c^{(n)} = i$ indicates that $G_{\text{dy}}^{(n)}$ is assigned to the structural proxy node $v^{(i)}$.
At initialization, we set $c_{\text{init}}^{(n)} = n$, yielding an one-to-one correspondence between dynamic Gaussians and structural proxy nodes. With this one-to-one correspondence, we initialize the spline control point set $\mathbf{P}^{(n)}$ by fitting it to the associated structural proxy node trajectory $\mathcal{Q}^{(c_\text{init}^{(n)})}$ as:
\begin{equation}
\mathbf{P}^{(n)}
=
\arg\min_{\mathbf{P}}
\sum_{t=1}^{N_f}
\left\|
\mathbf{q}_t^{(c_\text{init}^{(n)})}
-
S(t,\mathbf{P})
\right\|_2^2 ,
\label{eq:init_control_points}
\end{equation}
where $S(t,\mathbf{P})$ denotes the position of the cubic Hermite spline at time $t$ which is defined as:
\begin{equation}
S(t, \mathbf{P}) = \begin{bmatrix} t_r^3 & t_r^2 & t_r & 1 \end{bmatrix}
\begin{bmatrix}
\phantom{-}2 & -2 & \phantom{-}1 & \phantom{-}1 \\
-3 & \phantom{-}3 & -2 & -1 \\
\phantom{-}0 & \phantom{-}0 & \phantom{-}1 & \phantom{-}0 \\
\phantom{-}1 & \phantom{-}0 & \phantom{-}0 & \phantom{-}0
\end{bmatrix}
\begin{bmatrix}
\mathbf{p}_k \\ \mathbf{p}_{k+1} \\ \mathbf{m}_k \\ \mathbf{m}_{k+1}
\end{bmatrix},
\end{equation}
where $k=\lfloor t_s \rfloor$, $t_r = t_s - \lfloor t_s \rfloor$, $t_s = t(N_c-1)/(N_f-1)$, and $\mathbf{m}_k \approx (\mathbf{p}_{k+1} - \mathbf{p}_{k-1})/2$. The position of $G_{\text{dy}}^{(n)}$ at time $t$ is then defined as:
\begin{equation}
\bm{\mu}^{(n)}(t) = S(t,\mathbf{P}^{(n)}).
\end{equation}
During Gaussian densification, a newly split or cloned dynamic Gaussian $G_{\text{dy}}^{(n')}$ inherits the structural proxy node assignment of its source Gaussian $G_{\text{dy}}^{(n)}$ as $c^{(n')} = c^{(n)}$.
As a result, multiple dynamic Gaussians may share a specific structural proxy node, enabling newly generated Gaussians to inherit the structural guidance of the SPG $\mathcal{K}$. We thus define the Gaussian set associated with $v^{(i)}$ as:
\begin{equation}
\mathcal{G}_{\text{dy}}(v^{(i)}) = \{ G_{\text{dy}}^{(n)} \mid c^{(n)} = i \}.
\end{equation}
Notably, the association between each $G_{\text{dy}}^{(n)}$ and its assigned node $v^{(i)}$ is used only in training to enforce structural coherence in SIS learning. At inference time, Gaussian positions are computed solely by evaluating the learned spline functions, enabling fast rendering.

\subsection{Optimization}
\label{subsec:optimization}

To optimize the proposed WebSpline architecture, we adopt a two-stage optimization strategy: 
(i) a Structural Proxy Graph Foundation (SPGF) stage, in which we construct and optimize $\mathcal{K}$ to provide structural guidance, and 
(ii) a Structure-Informed Spline Optimization (SISO) stage, in which we learn the SIS representation under spatial and SPG-induced structural neighborhood constraints.

\textbf{SPGF Stage.}
In the SPGF stage, we optimize the SPG $\mathcal{K}$ to establish structural coherence over time among proxy nodes $\mathcal{V}$ representing dynamic regions. This coherent SPG enables stable SIS learning in the subsequent SISO stage. To this end, we introduce an orientation-similarity-based temporal rigidity (OTR) loss $\mathcal{L}_\text{OTR}$. $\mathcal{L}_\text{OTR}$ encourages temporal rigidity among neighboring nodes that exhibit similar rigid motion by considering both positional consistency and orientation similarity over time. For efficient optimization, we compute this temporal rigidity within a temporal window $\mathcal{W} = [t_1, t_2]$.
We first compute an orientation similarity score $s^{(i,j)}$ for each neighboring node pair $(v^{(i)}, v^{(j)})$ at the initial time $t_1$ of the temporal window as:
\begin{equation}
s^{(i,j)}
=
(\operatorname{tr}(
(\mathbf{O}_{t_1}^{(i)})^\top \mathbf{O}_{t_1}^{(j)}
)-1)/2,
\end{equation}
where $\operatorname{tr}(\cdot)$ denotes the trace operator.
We then compute the orientation similarity weight $w^{(i,j)}$ by applying a softmax to $s^{(i,j)}$ over the neighborhood $\mathcal{N}(i)$ of the structural proxy node $v^{(i)}$ as:
\begin{equation}
w^{(i,j)} = \exp(s^{(i,j)}) / \textstyle\sum_{j' \in \mathcal{N}(i)} \exp(s^{(i,j')}).
\end{equation}
Using $w^{(i,j)}$, $\mathcal{L}_\text{OTR}$ is computed within $\mathcal{W}$ as:
\begin{equation}
\begin{aligned}
    \mathcal{L}_\text{OTR} = 
    \sum_{t=t_1}^{t_2-\Delta}
    \sum_{i=1}^M
    \sum_{j\in\mathcal{N}(i)}
    w^{(i,j)} \Big(\left\| \mathbf{l}_t^{(i,j)} - \mathbf{l}_{t+\Delta}^{(i,j)} \right\|_2 + \Big|  d_t^{(i,j)} -  d_{t+\Delta}^{(i,j)}\Big| \Big), \\
    \text{where} \quad \mathbf{l}_t^{(i,j)} = (\mathbf{O}_t^{(j)})^\top(\mathbf{q}_t^{(i)}-\mathbf{q}_t^{(j)}), \quad     d_t^{(i,j)} = \| \mathbf{l}_t^{(i,j)} \|_2.
\end{aligned}
\label{eq:otr_loss}
\end{equation}
In Eq.~\ref{eq:otr_loss}, $\mathbf{l}_t^{(i,j)}$ denotes the relative offset from $v^{(j)}$ to $v^{(i)}$, expressed in the local coordinate system of node $v^{(j)}$ and $\Delta$ is a temporal offset over frames. Here, positional consistency refers to preserving the local relative offset $\mathbf{l}_t^{(i,j)}$ and distance $d_t^{(i,j)}$ between neighboring nodes across time, while orientation similarity is used to compute $w^{(i,j)}$. The overall objective for the SPGF stage is defined as:
\begin{equation}
\mathcal{L}_{\text{SPGF}} 
= 
\lambda_{\text{OTR}}\mathcal{L}_{\text{OTR}} 
+ 
\lambda_{\text{vel}}\mathcal{L}_{\text{vel}} 
+ 
\lambda_{\text{acc}}\mathcal{L}_{\text{acc}},
\label{eq:loss_spgf}
\end{equation}
where $\mathcal{L}_{\text{vel}}$ and $\mathcal{L}_{\text{acc}}$ denote the velocity and acceleration regularizations for the structural proxy nodes $\mathcal{V}$, respectively. Details are provided in the Appendix~\ref{subsec:vel_acc_reg}.

\textbf{SISO Stage.} In the SISO stage, we optimize the attributes of each dynamic Gaussian $G_{\text{dy}}^{(n)}$, including the learnable spline control point set $\mathbf{P}^{(n)}$ of the SIS representation, using the optimized SPG $\mathcal{K}$ as structural guidance.
To enhance the structural consistency of the SIS representation, we constrain the spline trajectories of dynamic Gaussians $\mathcal{G}_\text{dy}$ using two types of neighborhoods: spatial and structural.
For each dynamic Gaussian $G_{\text{dy}}^{(n)}$, we define its \textit{spatial neighborhood} $\mathcal{N}_{\text{sp}}(n)$ using the $K_\text{sp}$ nearest dynamic Gaussians measured from the first spline control point $\mathbf{p}_1^{(n)}$ as:
\begin{equation}
\mathcal{N}_{\text{sp}}(n)
= \{
m
\;|\;
m \neq n, \;
\operatorname{rank}_n
(
\| \mathbf{p}_1^{(n)} - \mathbf{p}_1^{(m)} \|_2
)
\le K_\text{sp}
\},
\label{eq:spatial_neighbor}
\end{equation}
where $\operatorname{rank}_n(\| \mathbf{p}_1^{(n)} - \mathbf{p}_1^{(m)} \|_2)$ is defined analogously to $\operatorname{rank}_i(\mathcal{D}_{ij})$ in Eq.~\eqref{eq:neighbor_and_edges}. We also define the \textit{structural neighborhood} $\mathcal{N}_{\text{st}}(n)$ of $G_{\text{dy}}^{(n)}$ based on the connectivity of its associated structural proxy node $v^{(c^{(n)})}$ with respect to the neighboring proxy nodes $\mathcal{N}(c^{(n)})$ in Eq.~\eqref{eq:neighbor_and_edges} as:
\begin{equation}
\mathcal{N}_{\text{st}}(n)
=
\bigcup_{j \in \mathcal{N}(c^{(n)})}
\{
m
\;|\;
G_{\text{dy}}^{(m)} \sim 
\mathcal{U}(\mathcal{G}_{\text{dy}}(v^{(j)}))
\},
\label{eq:structural_neighborhood}
\end{equation}
where $\mathcal{U}(\mathcal{G}_{\text{dy}}(v^{(j)}))$ denotes uniform sampling over the dynamic Gaussian set $\mathcal{G}_\text{dy}(v^{(j)})$ associated with $v^{(j)}$. We then define a spline temporal rigidity (STR) loss $\mathcal{L}_\text{STR}$ over ${N}_{\text{sp}}(n)$ and ${N}_{\text{st}}(n)$ as:
\begin{equation}
    \mathcal{L}_\text{STR} = 
    \underbrace{\alpha \cdot \sum_{k=1}^{N_c-\tau}
    \sum_{n=1}^{N_g} 
    \sum_{m \in \mathcal{N}_\text{sp}(n)} \Psi_k^{(n,m)}}_{\text{Spatial Spline Rigidity} \;(\mathcal{L}_\text{STR}^\text{sp})}
    +
    \underbrace{\beta \cdot \sum_{k=1}^{N_c-\tau}
    \sum_{n=1}^{N_g} 
    \sum_{m \in \mathcal{N}_\text{st}(n)} \Psi_k^{(n,m)}}_{\text{Structural Spline Rigidity} \;(\mathcal{L}_\text{STR}^\text{st})},
\label{eq:str_loss}
\end{equation}
where $\alpha$ and $\beta$ denote balancing coefficients for the spatial and structural spline rigidity terms, $\mathcal{L}_{\text{STR}}^{\text{sp}}$ and $\mathcal{L}_{\text{STR}}^{\text{st}}$, respectively. The spline rigidity $\Psi_k^{(n,m)}$ is defined as:
\begin{equation}
    \Psi_k^{(n,m)} = \| \mathbf{r}_k^{(n,m)} - \mathbf{r}_{k+\tau}^{(n,m)} \|_1 + | \rho_k^{(n,m)} - \rho_{k+\tau}^{(n,m)} |,
\end{equation}
where $\mathbf{r}_k^{(n,m)} = \mathbf{p}_k^{(m)} - \mathbf{p}_k^{(n)}$, $\rho_k^{(n,m)} = \| \mathbf{r}_k^{(n,m)} \|_2$ and $\tau$ denotes an index offset between control points within the same spline control point set. The two spline rigidity terms, $\mathcal{L}_{\text{STR}}^{\text{sp}}$ and $\mathcal{L}_{\text{STR}}^{\text{st}}$, provide complementary constraints for SIS optimization.
$\mathcal{L}_{\text{STR}}^{\text{sp}}$ preserves local geometric consistency among spatially nearby dynamic Gaussians, while $\mathcal{L}_{\text{STR}}^{\text{st}}$ promotes coherent motion among potentially distant but structurally related Gaussians associated with neighboring SPG nodes.
The resulting $\mathcal{L}_{\text{STR}}$ is combined with the RGB and depth losses, $\mathcal{L}_{\text{rgb}}$ and $\mathcal{L}_{\text{d}}$, to optimize the SIS representation for both structural coherence and reconstruction fidelity.
Here, $\mathcal{L}_{\text{rgb}}$ and $\mathcal{L}_{\text{d}}$ are L1 losses on the rendered frame and rendered depth, respectively.
The overall objective in the SISO stage is defined as:
\begin{equation}
\mathcal{L}_{\text{SISO}}
=
\lambda_{\text{STR}}\mathcal{L}_{\text{STR}}
+
\lambda_{\text{rgb}}\mathcal{L}_{\text{rgb}}
+
\lambda_{\text{d}}\mathcal{L}_{\text{d}}.
\label{eq:loss_siso}
\end{equation}
\section{Experiments}
\label{sec:experiments}
\begin{table}[t]
\centering
\caption{Quantitative comparisons of novel view synthesis on the iPhone \cite{gao2022monocular} and NVIDIA \cite{yoon2020dynamic} datasets. Best, second-best, and third-best results are highlighted in red, orange, and yellow, respectively. Rendering speed (FPS) and training time (Tr. h) are measured using a single NVIDIA RTX 3090 Ti GPU. Training times are measured on the NVIDIA dataset~\cite{yoon2020dynamic}.}
\label{tab:comparison_results}
\small
\renewcommand{\arraystretch}{1.15}
\resizebox{\textwidth}{!}{
\begin{tabular}{l |ccc | ccc | cc | c}
\toprule
\multicolumn{1}{c|}{\multirow{2}{*}{\textbf{Method}}} & \multicolumn{3}{c|}{\textbf{iPhone} \cite{gao2022monocular}} & \multicolumn{3}{c|}{\textbf{NVIDIA} \cite{yoon2020dynamic}} &  \multicolumn{2}{c|}{\textbf{FPS$\uparrow$}} & \multirow{2}{*}{\textbf{Tr. (h)$\downarrow$}} \\
\cline{2-4} \cline{5-7} \cline{8-9}
& \rule{0pt}{2.8ex}mPSNR$\uparrow$ & mSSIM$\uparrow$ & mLPIPS$\downarrow$ & PSNR$\uparrow$ & SSIM$\uparrow$ & LPIPS$\downarrow$  & iPhone & NVIDIA & \\
\midrule
T-NeRF (NeurIPS'22) \cite{gao2022monocular} & 16.96 & 0.577 &  0.379 & 18.33 & 0.436  & 0.511 & 0.43 & 0.28 & > 24 \\
HyperNeRF (ACM-TG'21) \cite{park2021hypernerf} & 16.81 & 0.569 & 0.332 & 17.60 & 0.377 & 0.367 & 0.27 & 0.33 & > 24 \\
NSFF (CVPR'21) \cite{li2021neural} & 15.46 & 0.551 & 0.396 & 24.33 & 0.751 & 0.199 & 0.30 & 0.38 & > 24 \\
4DGS (CVPR'24) \cite{liu2023robust} & 13.87 & 0.459 &  0.407 & 22.89 & 0.731 & 0.237 & 83  & 95 & \first{0.25} \\
RoDynRF (CVPR'23) \cite{liu2023robust} & 17.10 & 0.534 & 0.517 & 25.89 & \third{0.854} & 0.067 & 0.62 & 0.45 & > 24 \\
D-NPC (Eurographics’25) \cite{kappel2024d} & 16.41 & 0.582 & 0.319 & 25.64 &  0.845  & 0.109 & 65 & 70 & \second{0.27} \\
MarbleGS (SIGGRAPH'24) \cite{stearns2024dynamic} & 16.08 & 0.568 & 0.433 & 24.56 & 0.667 & 0.110 & 16 & 10 & 13 \\ 
HiMoR (CVPR'25) \cite{liang2025himor}    & 16.02 & 0.558 & 0.325 & 24.89 & 0.694 & 0.118 & \third{157} & 145 & 1 \\ 
MoDec-GS (CVPR'25) \cite{modecgs} & 14.65 & 0.320 & 0.461 & 23.94 & 0.628 & 0.132 & 15 & 10 & 1 \\ 
SoM (ICCV'25) \cite{som}     & 17.13 & 0.674 & 0.279 & 24.58 & 0.651 & 0.124 & 127 & \third{255} & 1 \\ 
SplineGS (CVPR'25) \cite{splinegs} & 18.70 & 0.663 & 0.325 & \second{27.12} & \second{0.872} & \second{0.052} & \first{288} & \first{400} & 2 \\
MoSca (CVPR'25) \cite{mosca}    & \third{19.54} & \third{0.719} & \third{0.247} & \third{26.48} & 0.848 & 0.072 & 45 & 40 & 0.8 \\
SE3BSplineGS (CVPR'26) \cite{se3bsplinegs}    & 19.36 & 0.705 & 0.290 & 26.15 & 0.845 & 0.085 & 64 & 56 & 2 \\
WorldTree (ICLR'26) \cite{worldtree}    & \second{19.75} &\second{0.728} & \second{0.240} & 26.28 & 0.845 & \third{0.065} & 27 & 47 & 1 \\
\midrule
\textbf{WebSpline(Ours)} & \first{20.11} & \first{0.738} & \first{0.222} & \first{27.73} & \first{0.881} & \first{0.050} & \second{278} & \second{390} & \third{0.5} \\
\bottomrule
\end{tabular}
}
\vspace{-0.3cm}
\end{table}
\begin{figure}[t]
    \centering
    \includegraphics[width=\linewidth,keepaspectratio]{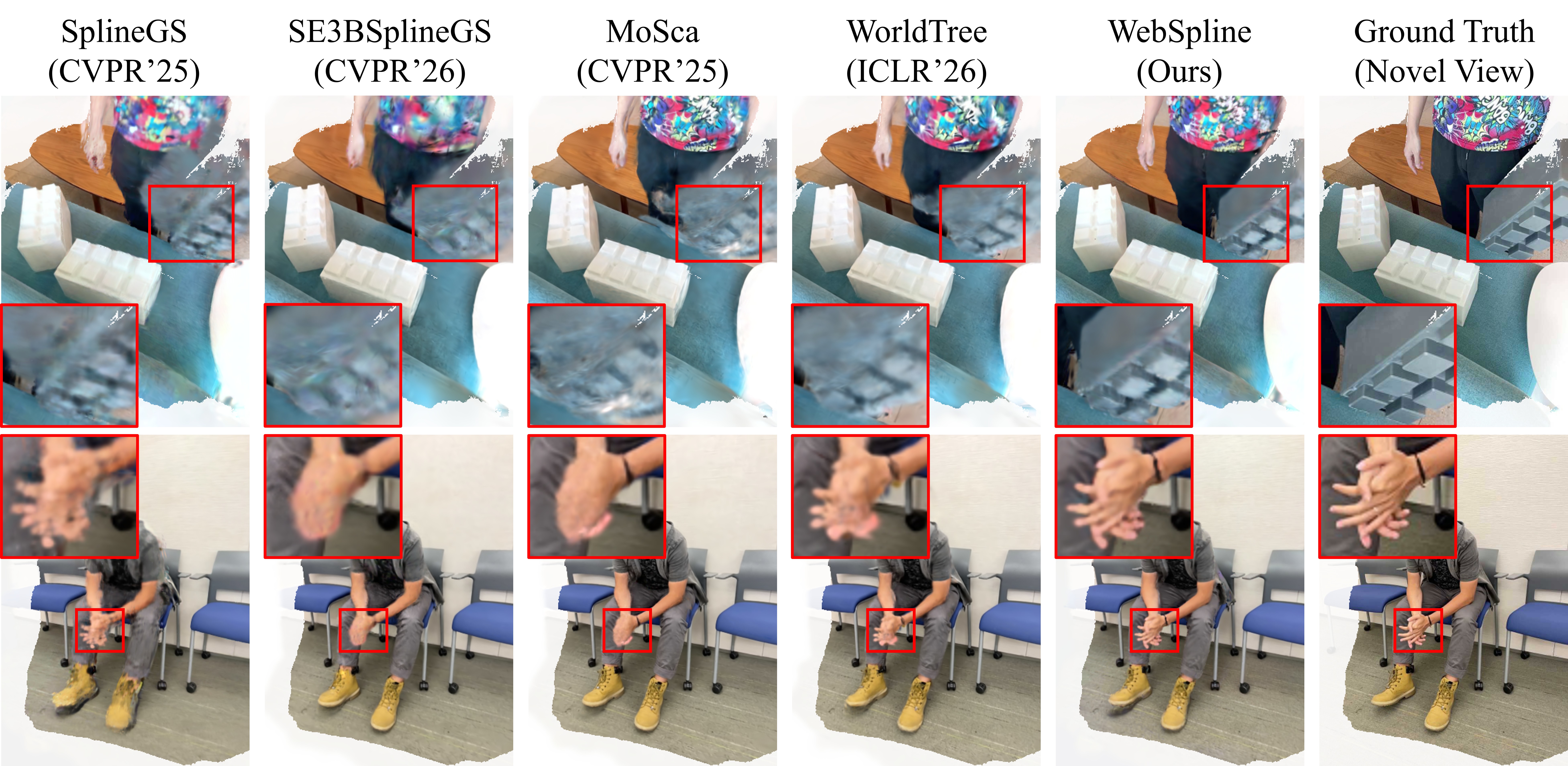}
    \vspace{-0.5cm}
    \caption{\textbf{Visual comparisons for novel view synthesis on the iPhone dataset \cite{gao2022monocular}.}}
    \label{fig:qualitative_iphone}
    \vspace{-0.6cm}
\end{figure}

\begin{figure}[t]
    \centering
    \includegraphics[width=\linewidth,keepaspectratio]{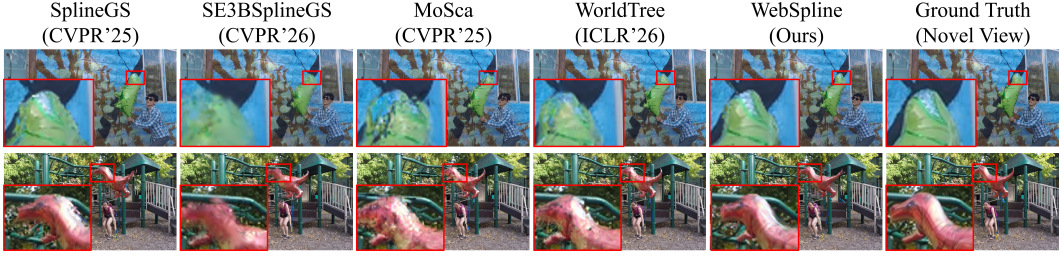}
    \vspace{-0.55cm}
    \caption{\textbf{Visual comparisons for novel view synthesis on the NVIDIA dataset \cite{yoon2020dynamic}.}}
    \label{fig:qualitative_nvidia}
    \vspace{-0.3cm}
\end{figure}

\begin{figure}
    \centering
    \includegraphics[width=\linewidth,keepaspectratio]{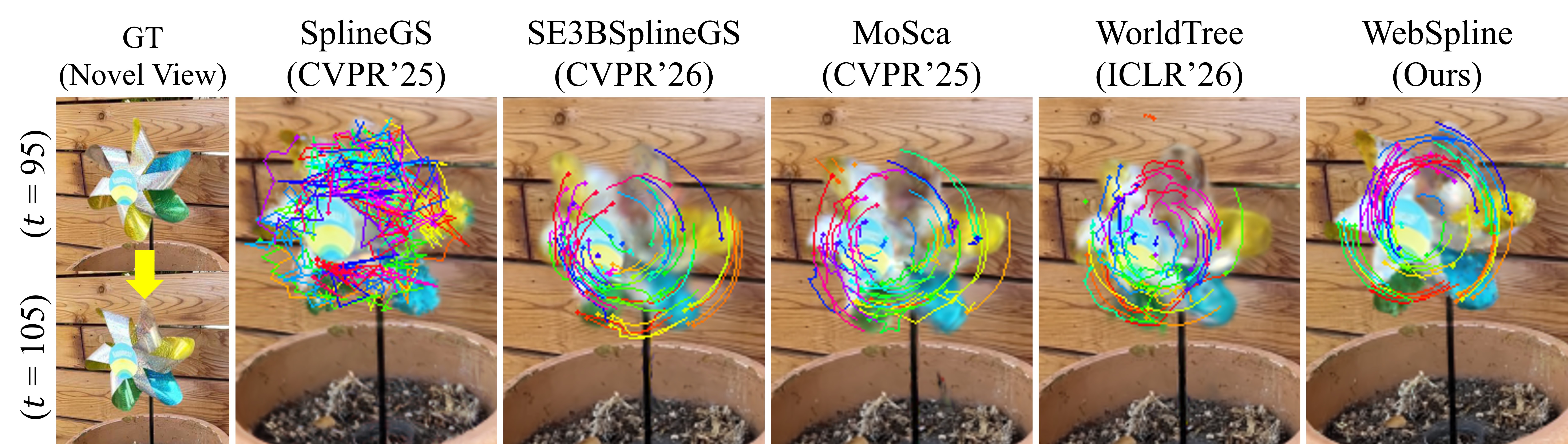}
    \vspace{-0.55cm}
    \caption{\textbf{Visual comparisons of dynamic Gaussian trajectories on the iPhone dataset~\cite{gao2022monocular}.}}
    \label{fig:trajectory_viz}
    \vspace{-0.6cm}
\end{figure}

\subsection{Experimental Setup}

\textbf{Implementation Details.} We refer readers to the Appendix~\ref{sec:implementation} for the details of our implementation.

\textbf{Datasets.} We evaluate WebSpline on two widely used monocular video benchmarks: the iPhone~\cite{gao2022monocular} and NVIDIA~\cite{yoon2020dynamic} datasets. The iPhone dataset consists of 7 casually captured scenes featuring complex camera and object motions. The NVIDIA dataset includes 7 scenes with diverse motion patterns, originally recorded with a 12-camera rig. Following the protocol of RoDynRF~\cite{liu2023robust}, we simulate a monocular video sequence by selecting one camera view per time step.

\textbf{Metrics.}
For the iPhone dataset~\cite{gao2022monocular}, we follow DyCheck~\cite{gao2022monocular} and report co-visibility-masked mPSNR, mSSIM, and mLPIPS to exclude regions unobserved during training.
For the NVIDIA dataset~\cite{yoon2020dynamic}, we report PSNR, SSIM, and LPIPS.
We also report FPS to assess rendering efficiency.

\subsection{Comparison with State-of-the-Art Methods}
\label{sec:experiments_result}

\textbf{Novel View Synthesis.} Table~\ref{tab:comparison_results} presents quantitative comparisons for dynamic novel view synthesis between our WebSpline and existing methods~\cite{gao2022monocular, park2021hypernerf, li2021neural, 4dgs, liu2023robust, kappel2024d, stearns2024dynamic, liang2025himor, modecgs, som, splinegs, mosca, se3bsplinegs, worldtree} on the iPhone~\cite{gao2022monocular} and NVIDIA~\cite{yoon2020dynamic} datasets. Since SE3BSplineGS~\cite{se3bsplinegs} only reported performance for a subset of the iPhone dataset ~\cite{gao2022monocular}, we conducted an evaluation across all seven scenes using their official code and evaluated it under the same setting. As shown in Table~\ref{tab:comparison_results}, WebSpline consistently outperforms all the SOTA methods across all metrics on both datasets. The scaffold-based methods~\cite{mosca, se3bsplinegs, worldtree} show lower perceptual quality than our WebSpline on both datasets, as reflected by mLPIPS and LPIPS, indicating the limitation of their reliance on motion blending. SplineGS~\cite{splinegs} achieves the second-best results on the NVIDIA dataset~\cite{yoon2020dynamic}, but its performance degrades on the longer and more challenging iPhone~\cite{gao2022monocular} sequences, implying the instability of its unconstrained motion representation. Notably, WebSpline achieves 278 FPS on the iPhone dataset, rendering more than \textbf{10$\times$ faster} than WorldTree~\cite{worldtree} that is the second-best method in rendering quality but achieves only 27 FPS in rendering speed.
While WorldTree~\cite{worldtree} computes each Gaussian motion by blending neighboring scaffold-node motions using dual quaternion blending (DQB)~\cite{kavan2007skinning}, our SIS representation directly evaluates spline-based Gaussian motion, avoiding this blending overhead and enabling substantially faster rendering. Figs.~\ref{fig:qualitative_iphone} and~\ref{fig:qualitative_nvidia} show visual comparisons between WebSpline and SOTA methods~\cite{splinegs, se3bsplinegs, mosca, worldtree} on the iPhone~\cite{gao2022monocular} and NVIDIA~\cite{yoon2020dynamic} datasets, respectively. As shown, WebSpline produces novel view renderings with clearer boundaries and sharper details of dynamic objects than the SOTA methods~\cite{splinegs, se3bsplinegs, mosca,  worldtree}. More results are provided in the Appendix~\ref{sec:additional_visual_results}.

\textbf{Gaussian Trajectory Visualization.} To further evaluate WebSpline's motion modeling, Fig.~\ref{fig:trajectory_viz} visualizes dynamic Gaussian trajectories projected onto the 2D image plane and compares them with SOTA methods~\cite{splinegs, se3bsplinegs, mosca, worldtree}. SplineGS~\cite{splinegs} fails to accurately model the trajectories of moving dynamic regions, resulting in noisy Gaussian motions. The scaffold-based methods~\cite{mosca, se3bsplinegs, worldtree} tend to produce jittering trajectories, indicating that motion blending across neighboring scaffold nodes may lead to temporally unstable Gaussian motion. In contrast, our WebSpline produces clear trajectories, showing both structurally coherent and temporally smooth Gaussian motion representation.

\subsection{Ablation Study}
\label{sec:ablation}

\begin{figure}
    \centering
    \includegraphics[width=\linewidth,keepaspectratio]{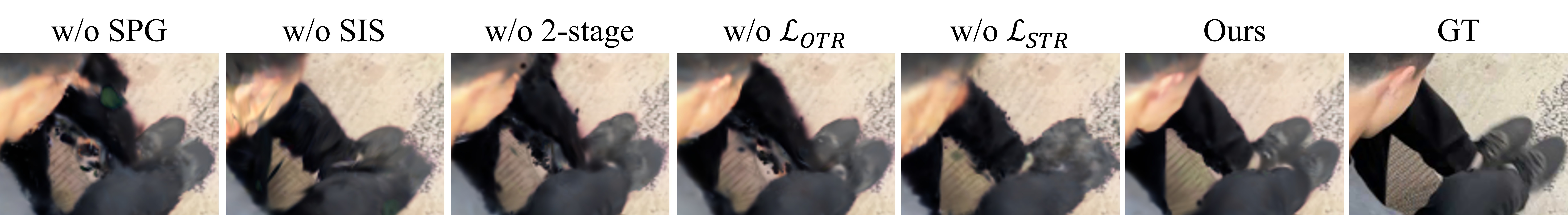}
    \vspace{-0.5cm}
    \caption{\textbf{Visual results of ablation study on the iPhone dataset~\cite{gao2022monocular}.}}
    \label{fig:ablation}
    \vspace{-0.6cm}
\end{figure}

To validate the effectiveness of WebSpline's framework design and loss functions, 
Table~\ref{tab:ablation_study} and Fig.~\ref{fig:ablation} present ablation results on the iPhone dataset~\cite{gao2022monocular}.

\textbf{Framework Design.} Table~\ref{tab:ablation_study}-(a) reports ablation results for the framework design of WebSpline, including the SPG, SIS, and two-stage optimization strategy.
For `w/o SPG', we remove the SPG and initialize the spline-based dynamic Gaussian trajectories directly from 2D point tracks~\cite{karaev2025cotracker3}. Without the SPG, dynamic Gaussian trajectories are optimized independently, making them more prone to noise and incoherence. This leads to degraded performance, showing the importance of the SPG for coherent Gaussian motion.
For `w/o SIS', we remove the SIS representation and directly use the structural proxy node trajectories obtained after the SPGF stage to represent dynamic Gaussian motion. The resulting performance drop shows that SIS plays a crucial role in providing sufficient motion expressiveness for fine-grained dynamic Gaussian trajectories.
For `w/o 2-stage', we remove the SPGF stage and use the SPG initialized from 2D point tracks without pre-optimization. The SIS is initialized from this unoptimized SPG, and the SPG and SIS are then jointly optimized in a single stage.
The resulting degraded performance indicates that a pre-optimized SPG provides more reliable structural guidance for stable SIS learning, highlighting the rationale behind our two-stage optimization strategy. These trends are also consistent with the qualitative results in Fig.~\ref{fig:ablation}, where all three ablated variants exhibit noisier and less coherent motion than the full WebSpline.

\begin{table}[t]
\centering
\caption{\textbf{Ablation studies.} We ablate the key framework design components and loss functions of WebSpline on the iPhone dataset~\cite{gao2022monocular} under the same setting as Table~\ref{tab:comparison_results}.}
\vspace{-0.2cm}
\begin{minipage}[t]{0.49\linewidth}
\centering
\captionsetup{justification=centering}
\caption*{(a) Framework Design}
\scalebox{0.88}{
\begin{tabular}{l|ccc}
\toprule
\multicolumn{1}{c|}{Method} & mPSNR$\uparrow$ & mSSIM$\uparrow$ & mLPIPS$\downarrow$ \\
\midrule
w/o SPG & 19.68 & 0.728 & 0.236 \\
w/o SIS & 19.62 & 0.724 & 0.239 \\
w/o 2-stage & 19.75 & 0.730 & 0.230 \\
\midrule
Ours & \first{20.11} & \first{0.738} & \first{0.222} \\
\bottomrule
\end{tabular}
}
\end{minipage}
\hfill
\begin{minipage}[t]{0.49\linewidth}
\centering
\captionsetup{justification=centering}
\caption*{(b) Loss Functions}
\scalebox{0.77}{
\begin{tabular}{l|ccc}
\toprule
\multicolumn{1}{c|}{Method} & mPSNR$\uparrow$ & mSSIM$\uparrow$ & mLPIPS$\downarrow$ \\
\midrule
w/o $\mathcal{L}_{\text{OTR}}$ & 19.79 & 0.728 & 0.232 \\
w/o $\mathcal{L}_{\text{STR}}$ & 19.76 & 0.729 & 0.245 \\
w/o $\mathcal{L}_{\text{STR}}^{\text{sp}}$ & 20.04 & 0.738 & 0.226 \\
w/o $\mathcal{L}_{\text{STR}}^{\text{st}}$ & 20.01 & 0.737 & 0.227 \\
\midrule
Ours & \first{20.11} & \first{0.738} & \first{0.222} \\
\bottomrule
\end{tabular}
}
\end{minipage}
\label{tab:ablation_study}
\vspace{-0.3cm}
\end{table}

\textbf{Loss Functions.}
Table~\ref{tab:ablation_study}-(b) shows the effectiveness of $\mathcal{L}_\text{OTR}$ and $\mathcal{L}_\text{STR}$, used in the SPGF and SISO stages, respectively. Without $\mathcal{L}_\text{OTR}$, the SPG fails to maintain structurally coherent node trajectories over time, weakening the structural guidance for SIS learning. Without $\mathcal{L}_\text{STR}$, neighboring Gaussians lose temporal rigidity, resulting in less coherent spline trajectories. This indicates that explicit rigidity constraints are necessary beyond mere SPG initialization. Fig.~\ref{fig:ablation} further shows that removing either $\mathcal{L}_{\text{OTR}}$ or $\mathcal{L}_{\text{STR}}$ leads to noisier renderings, consistent with the quantitative degradation. Furthermore, as shown in Table~\ref{tab:ablation_study}-(b), we ablate the two spline rigidity terms $\mathcal{L}_\text{STR}^\text{sp}$ and $\mathcal{L}_\text{STR}^\text{st}$ in $\mathcal{L}_\text{STR}$ to analyze their individual contributions. Ablating either $\mathcal{L}_\text{STR}^\text{sp}$ or $\mathcal{L}_\text{STR}^\text{st}$ degrades performance, while using both terms achieves the best results.
This demonstrates their complementary roles in achieving local geometric consistency and SPG-induced structural coherence, respectively.

\begin{wrapfigure}{r}{0.48\linewidth}
    \centering
    \vspace{-0.5cm}
    \includegraphics[width=\linewidth,keepaspectratio]{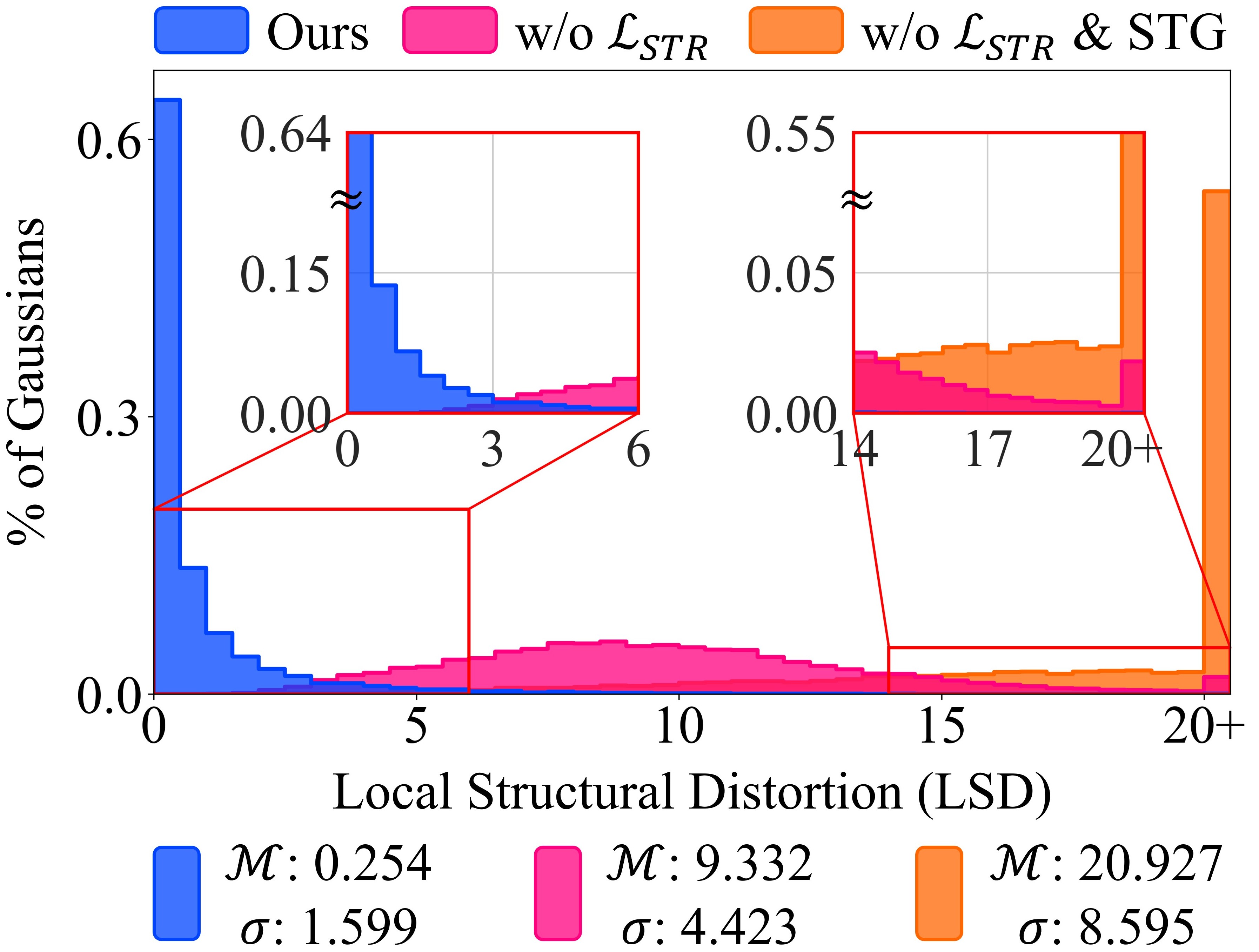}
    \caption{\textbf{Distributions of LSD Values.}
    We use the \textit{paper-windmill} scene of the iPhone dataset~\cite{gao2022monocular}. `20+' indicates all LSD values above 20 are accumulated at the LSD value of 20.}
    \label{fig:analysis}
    \vspace{-0.5cm}
\end{wrapfigure}

\textbf{Analysis of Local Structural Distortion (LSD).}
To further analyze the structural consistency of WebSpline, Fig.~\ref{fig:analysis} visualizes the local structural distortion (LSD) distribution across dynamic Gaussians. LSD measures the temporal variation of relative distances among neighboring Gaussians. Specifically, for each $G_{\text{dy}}^{(n)}$, we identify its $K=8$ nearest neighbors and compute the mean temporal variance of its distances to these neighbors. We compare WebSpline against two variants: (i) `w/o $\mathcal{L}_{\text{STR}}$' and (ii) `w/o $\mathcal{L}_{\text{STR}}$ \& STG'. In Fig.~\ref{fig:analysis}, we provide the median ($\mathcal{M}$) and standard deviation ($\sigma$) of each distribution. For `w/o $\mathcal{L}_{\text{STR}}$ \& STG' (orange), the LSD's $\mathcal{M}$ reaches 20.927, indicating severe local structural distortion caused by unconstrained Gaussian trajectories. While `w/o $\mathcal{L}_{\text{STR}}$' (pink) lowers the LSD's $\mathcal{M}$, it still exhibits high variance ($\sigma=4.423$), showing its insufficiency in consistently constraining Gaussian trajectories. In contrast, the LSD values of our full model (blue) are concentrated in the low-value range, confirming that WebSpline enables dynamic Gaussians to move in a structurally stable manner with respect to their neighboring Gaussians.

\section{Conclusion}
\label{sec:conclusion}

We present WebSpline, a dynamic 3D Gaussian framework for structurally coherent and high-fidelity reconstruction from monocular videos. At its core, WebSpline explicitly couples a Structure-Informed Spline (SIS) representation with a Structural Proxy Graph (SPG), achieving a highly expressive yet robust motion model that captures complex, fine-grained Gaussian dynamics while strictly preventing structural degradation. With two-stage optimization and SIS-only inference, WebSpline achieves \textbf{state-of-the-art rendering quality} on the iPhone and NVIDIA datasets while enabling \textbf{over $10\times$ faster} rendering than WorldTree, the second-best model on the iPhone dataset.

{
    \small
    \bibliographystyle{plainnat}
    \bibliography{main}

@String(CVPR= {IEEE Conf. Comput. Vis. Pattern Recog.})

@String(ICCV= {Int. Conf. Comput. Vis.})

@String(TOG= {ACM Trans. Graph.})

@String(AAAI = {AAAI})

@String(CVPR  = {CVPR})

@String(ICCV  = {ICCV})

@String(TOG   = {ACM TOG})

@misc{nerf,
      title={NeRF: Representing Scenes as Neural Radiance Fields for View Synthesis}, 
      author={Ben Mildenhall and Pratul P. Srinivasan and Matthew Tancik and Jonathan T. Barron and Ravi Ramamoorthi and Ren Ng},
      year={2020},
      eprint={2003.08934},
      archivePrefix={arXiv},
      primaryClass={cs.CV},
      url={https://arxiv.org/abs/2003.08934}, 
}

@misc{mipnerf,
      title={Mip-NeRF: A Multiscale Representation for Anti-Aliasing Neural Radiance Fields}, 
      author={Jonathan T. Barron and Ben Mildenhall and Matthew Tancik and Peter Hedman and Ricardo Martin-Brualla and Pratul P. Srinivasan},
      year={2021},
      eprint={2103.13415},
      archivePrefix={arXiv},
      primaryClass={cs.CV},
      url={https://arxiv.org/abs/2103.13415}, 
}

@misc{mipnerf360,
      title={Mip-NeRF 360: Unbounded Anti-Aliased Neural Radiance Fields}, 
      author={Jonathan T. Barron and Ben Mildenhall and Dor Verbin and Pratul P. Srinivasan and Peter Hedman},
      year={2022},
      eprint={2111.12077},
      archivePrefix={arXiv},
      primaryClass={cs.CV},
      url={https://arxiv.org/abs/2111.12077}, 
}

@misc{3dgs,
      title={3D Gaussian Splatting for Real-Time Radiance Field Rendering}, 
      author={Bernhard Kerbl and Georgios Kopanas and Thomas Leimkühler and George Drettakis},
      year={2023},
      eprint={2308.04079},
      archivePrefix={arXiv},
      primaryClass={cs.GR},
      url={https://arxiv.org/abs/2308.04079}, 
}

@misc{dyn3dgs,
      title={Dynamic 3D Gaussians: Tracking by Persistent Dynamic View Synthesis}, 
      author={Jonathon Luiten and Georgios Kopanas and Bastian Leibe and Deva Ramanan},
      year={2023},
      eprint={2308.09713},
      archivePrefix={arXiv},
      primaryClass={cs.CV},
      url={https://arxiv.org/abs/2308.09713}, 
}

@misc{4dgs,
      title={4D Gaussian Splatting for Real-Time Dynamic Scene Rendering}, 
      author={Guanjun Wu and Taoran Yi and Jiemin Fang and Lingxi Xie and Xiaopeng Zhang and Wei Wei and Wenyu Liu and Qi Tian and Xinggang Wang},
      year={2024},
      eprint={2310.08528},
      archivePrefix={arXiv},
      primaryClass={cs.CV},
      url={https://arxiv.org/abs/2310.08528}, 
}

@article{deformable3dgs,
    title={Deformable 3D Gaussians for High-Fidelity Monocular Dynamic Scene Reconstruction},
    author={Yang, Ziyi and Gao, Xinyu and Zhou, Wen and Jiao, Shaohui and Zhang, Yuqing and Jin, Xiaogang},
    journal={arXiv preprint arXiv:2309.13101},
    year={2023}
}

@article{mosca,
  title={MoSca: Dynamic Gaussian Fusion from Casual Videos via 4D Motion Scaffolds},
  author={Lei, Jiahui and Weng, Yijia and Harley, Adam and Guibas, Leonidas and Daniilidis, Kostas},
  journal={arXiv preprint arXiv:2405.17421},
  year={2024}
}

@inproceedings{som,
  title     = {Shape of Motion: 4D Reconstruction from a Single Video},
  author    = {Wang, Qianqian and Ye, Vickie and Gao, Hang and Zeng, Weijia and Austin, Jake and Li, Zhengqi and Kanazawa, Angjoo},
  booktitle   = {International Conference on Computer Vision (ICCV)},
  year      = {2025}
}

@inproceedings{st4dgs,
author = {Li, Deqi and Huang, Shi-Sheng and Lu, Zhiyuan and Duan, Xinran and Huang, Hua},
title = {ST-4DGS: Spatial-Temporally Consistent 4D Gaussian Splatting for Efficient Dynamic Scene Rendering},
year = {2024},
isbn = {9798400705250},
publisher = {Association for Computing Machinery},
address = {New York, NY, USA},
url = {https://doi.org/10.1145/3641519.3657520},
doi = {10.1145/3641519.3657520},
booktitle = {ACM SIGGRAPH 2024 Conference Papers},
articleno = {83},
numpages = {11},
location = {Denver, CO, USA},
series = {SIGGRAPH '24}
}

@InProceedings{modecgs,
  title={MoDec-GS: Global-to-Local Motion Decomposition and Temporal Interval Adjustment for Compact Dynamic 3D Gaussian Splatting}, 
  author={Sangwoon Kwak and Joonsoo Kim and Jun Young Jeong and Won-Sik Cheong and Jihyong Oh and Munchurl Kim},
  booktitle = {Proceedings of the IEEE/CVF Conference on Computer Vision and Pattern Recognition (CVPR)},
  year={2025},
}

@InProceedings{splinegs,
      author    = {Park, Jongmin and Bui, Minh-Quan Viet and Bello, Juan Luis Gonzalez and Moon, Jaeho and Oh, Jihyong and Kim, Munchurl},
      title     = {SplineGS: Robust Motion-Adaptive Spline for Real-Time Dynamic 3D Gaussians from Monocular Video},
      booktitle = {Proceedings of the Computer Vision and Pattern Recognition Conference (CVPR)},
      month     = {June},
      year      = {2025},
      pages     = {26866-26875}
  }

@inproceedings{
worldtree,
title={WorldTree: Towards 4D Dynamic Worlds from Monocular Video using Tree-Chains},
author={Qisen Wang and Yifan Zhao and Jia Li},
booktitle={The Fourteenth International Conference on Learning Representations},
year={2026},
url={https://openreview.net/forum?id=mVo6cyFR6C}
}

@article{karaev2023cotracker,
  title={Cotracker: It is better to track together},
  author={Karaev, Nikita and Rocco, Ignacio and Graham, Benjamin and Neverova, Natalia and Vedaldi, Andrea and Rupprecht, Christian},
  journal={arXiv},
  year={2023}
}

@inproceedings{piccinelli2024unidepth,
  title={UniDepth: Universal Monocular Metric Depth Estimation},
  author={Piccinelli, Luigi and Yang, Yung-Hsu and Sakaridis, Christos and Segu, Mattia and Li, Siyuan and Van Gool, Luc and Yu, Fisher},
  booktitle={CVPR},
  year={2024}
}

@inproceedings{liu2023robust,
  author    = {Liu, Yu-Lun and Gao, Chen and Meuleman, Andreas and Tseng, Hung-Yu and Saraf, Ayush and Kim, Changil and Chuang, Yung-Yu and Kopf, Johannes and Huang, Jia-Bin},
  title     = {Robust Dynamic Radiance Fields},
  booktitle = {CVPR},
  year      = {2023}
}

@inproceedings{Li_STG_2024_CVPR,
    author    = {Li, Zhan and Chen, Zhang and Li, Zhong and Xu, Yi},
    title     = {Spacetime Gaussian Feature Splatting for Real-Time Dynamic View Synthesis},
    booktitle = {CVPR},
    year      = {2024}
}

@inproceedings{yoon2020dynamic,
title={Novel View Synthesis of Dynamic Scenes with Globally Coherent Depths from a Monocular Camera},
author={Yoon, Jae Shin and Kim, Kihwan and Gallo, Orazio and Park, Hyun Soo and Kautz, Jan},
booktitle={CVPR},
year={2020}
}

@article{park2021hypernerf,
  author = {Park, Keunhong and Sinha, Utkarsh and Hedman, Peter and Barron, Jonathan T. and Bouaziz, Sofien and Goldman, Dan B and Martin-Brualla, Ricardo and Seitz, Steven M.},
  title = {HyperNeRF: A Higher-Dimensional Representation for Topologically Varying Neural Radiance Fields},
  journal = {ACM Trans. Graph.},
  year = {2021},
}

@inproceedings{li2021neural,
  title={Neural scene flow fields for space-time view synthesis of dynamic scenes},
  author={Li, Zhengqi and Niklaus, Simon and Snavely, Noah and Wang, Oliver},
  booktitle={CVPR},
  year={2021}
}

@article{huang2023sc,
  title={SC-GS: Sparse-Controlled Gaussian Splatting for Editable Dynamic Scenes},
  author={Huang, Yi-Hua and Sun, Yang-Tian and Yang, Ziyi and Lyu, Xiaoyang and Cao, Yan-Pei and Qi, Xiaojuan},
  journal={CVPR},
  year={2024}
}

@book{ahlberg2016theory,
  title={The Theory of Splines and Their Applications: Mathematics in Science and Engineering: A Series of Monographs and Textbooks, Vol. 38},
  author={Ahlberg, J Harold and Nilson, Edwin Norman and Walsh, Joseph Leonard},
  volume={38},
  year={2016},
  publisher={Elsevier}
}

@article{de1978practical,
  title={A practical guide to splines},
  author={De Boor, C},
  journal={Springer-Verlag google schola},
  year={1978}
}

@article{liang2023gaufre,
  title={Gaufre: Gaussian deformation fields for real-time dynamic novel view synthesis},
  author={Liang, Yiqing and Khan, Numair and Li, Zhengqin and Nguyen-Phuoc, Thu and Lanman, Douglas and Tompkin, James and Xiao, Lei},
  journal={arXiv},
  year={2023}
}

@inproceedings{lee2024fully,
  title={Fully Explicit Dynamic Gaussian Splatting},
  author={Lee, Junoh and Won, Chang-Yeon and Jung, Hyunjun and Bae, Inhwan and Jeon, Hae-Gon},
    booktitle={NeurIPS},
    year={2024},
}

@article{wan2024superpoint,
  title={Superpoint gaussian splatting for real-time high-fidelity dynamic scene reconstruction},
  author={Wan, Diwen and Lu, Ruijie and Zeng, Gang},
  journal={arXiv preprint arXiv:2406.03697},
  year={2024}
}

@inproceedings{bae2024per,
  title={Per-gaussian embedding-based deformation for deformable 3d gaussian splatting},
  author={Bae, Jeongmin and Kim, Seoha and Yun, Youngsik and Lee, Hahyun and Bang, Gun and Uh, Youngjung},
  booktitle={European Conference on Computer Vision},
  pages={321--335},
  year={2024},
  organization={Springer}
}

@article{kappel2024d,
  title={D-npc: Dynamic neural point clouds for non-rigid view synthesis from monocular video},
  author={Kappel, Moritz and Hahlbohm, Florian and Scholz, Timon and Castillo, Susana and Theobalt, Christian and Eisemann, Martin and Golyanik, Vladislav and Magnor, Marcus},
  journal={arXiv preprint arXiv:2406.10078},
  year={2024}
}

@inproceedings{chen2022tensorf,
  title={Tensorf: Tensorial radiance fields},
  author={Chen, Anpei and Xu, Zexiang and Geiger, Andreas and Yu, Jingyi and Su, Hao},
  booktitle={European conference on computer vision},
  pages={333--350},
  year={2022},
  organization={Springer}
}

@inproceedings{sun2022direct,
  title={Direct voxel grid optimization: Super-fast convergence for radiance fields reconstruction},
  author={Sun, Cheng and Sun, Min and Chen, Hwann-Tzong},
  booktitle={Proceedings of the IEEE/CVF conference on computer vision and pattern recognition},
  pages={5459--5469},
  year={2022}
}

@inproceedings{wang2022fourier,
  title={Fourier plenoctrees for dynamic radiance field rendering in real-time},
  author={Wang, Liao and Zhang, Jiakai and Liu, Xinhang and Zhao, Fuqiang and Zhang, Yanshun and Zhang, Yingliang and Wu, Minye and Yu, Jingyi and Xu, Lan},
  booktitle={Proceedings of the IEEE/CVF Conference on Computer Vision and Pattern Recognition},
  pages={13524--13534},
  year={2022}
}

@inproceedings{fridovich2022plenoxels,
  title={Plenoxels: Radiance fields without neural networks},
  author={Fridovich-Keil, Sara and Yu, Alex and Tancik, Matthew and Chen, Qinhong and Recht, Benjamin and Kanazawa, Angjoo},
  booktitle={Proceedings of the IEEE/CVF conference on computer vision and pattern recognition},
  pages={5501--5510},
  year={2022}
}

@article{muller2022instant,
  title={Instant neural graphics primitives with a multiresolution hash encoding},
  author={M{\"u}ller, Thomas and Evans, Alex and Schied, Christoph and Keller, Alexander},
  journal={ACM transactions on graphics (TOG)},
  volume={41},
  number={4},
  pages={1--15},
  year={2022},
  publisher={ACM New York, NY, USA}
}

@inproceedings{yu2021plenoctrees,
  title={Plenoctrees for real-time rendering of neural radiance fields},
  author={Yu, Alex and Li, Ruilong and Tancik, Matthew and Li, Hao and Ng, Ren and Kanazawa, Angjoo},
  booktitle={Proceedings of the IEEE/CVF international conference on computer vision},
  pages={5752--5761},
  year={2021}
}

@article{gao2022monocular,
 title={Monocular dynamic view synthesis: A reality check},
 author={Gao, Hang and Li, Ruilong and Tulsiani, Shubham and Russell, Bryan and Kanazawa, Angjoo},
 journal={Advances in Neural Information Processing Systems},
 volume={35},
 pages={33768--33780},
 year={2022}
}

@inproceedings{liang2025himor,
  title={Himor: Monocular deformable gaussian reconstruction with hierarchical motion representation},
  author={Liang, Yiming and Xu, Tianhan and Kikuchi, Yuta},
  booktitle={Proceedings of the IEEE/CVF Conference on Computer Vision and Pattern Recognition},
  pages={886--895},
  year={2025}
}

@inproceedings{stearns2024dynamic,
  title={Dynamic gaussian marbles for novel view synthesis of casual monocular videos},
  author={Stearns, Colton and Harley, Adam and Uy, Mikaela and Dubost, Florian and Tombari, Federico and Wetzstein, Gordon and Guibas, Leonidas},
  booktitle={SIGGRAPH Asia 2024 Conference Papers},
  pages={1--11},
  year={2024}
}

@article{se3bsplinegs,
  title={Learning Explicit Continuous Motion Representation for Dynamic Gaussian Splatting from Monocular Videos},
  author={Zhang, Xuankai and Xiao, Junjin and Huang, Shangwei and Zheng, Wei-shi and Zhang, Qing},
  journal={arXiv preprint arXiv:2603.25058},
  year={2026}
}

@inproceedings{teed2020raft,
  title={Raft: Recurrent all-pairs field transforms for optical flow},
  author={Teed, Zachary and Deng, Jia},
  booktitle={European conference on computer vision},
  pages={402--419},
  year={2020},
  organization={Springer}
}

@inproceedings{karaev2025cotracker3,
  title={Cotracker3: Simpler and better point tracking by pseudo-labelling real videos},
  author={Karaev, Nikita and Makarov, Yuri and Wang, Jianyuan and Neverova, Natalia and Vedaldi, Andrea and Rupprecht, Christian},
  booktitle={Proceedings of the IEEE/CVF International Conference on Computer Vision},
  pages={6013--6022},
  year={2025}
}

@INPROCEEDINGS{sam,
  author={Kirillov, Alexander and Mintun, Eric and Ravi, Nikhila and Mao, Hanzi and Rolland, Chloe and Gustafson, Laura and Xiao, Tete and Whitehead, Spencer and Berg, Alexander C. and Lo, Wan-Yen and Dollár, Piotr and Girshick, Ross},
  booktitle={2023 IEEE/CVF International Conference on Computer Vision (ICCV)}, 
  title={Segment Anything}, 
  year={2023},
  volume={},
  number={},
  pages={3992-4003},
  doi={10.1109/ICCV51070.2023.00371}}

@inproceedings{depthanything,
  title={Depth Anything: Unleashing the Power of Large-Scale Unlabeled Data},
  author={Yang, Lihe and Kang, Bingyi and Huang, Zilong and Xu, Xiaogang and Feng, Jiashi and Zhao, Hengshuang},
  booktitle={CVPR},
  year={2024}
}

@inproceedings{kavan2007skinning,
  title={Skinning with dual quaternions},
  author={Kavan, Ladislav and Collins, Steven and {\v{Z}}{\'a}ra, Ji{\v{r}}{\'\i} and O'Sullivan, Carol},
  booktitle={Proceedings of the 2007 symposium on Interactive 3D graphics and games},
  pages={39--46},
  year={2007}
}

@inproceedings{qingming2025modgs,
  title={Modgs: Dynamic gaussian splatting from casually-captured monocular videos with depth priors},
  author={Qingming, LIU and Liu, Yuan and Wang, Jiepeng and Lyu, Xianqiang and Wang, Peng and Wang, Wenping and Hou, Junhui},
  booktitle={The Thirteenth International Conference on Learning Representations},
  year={2025}
}

@inproceedings{bui2026mobgs,
  title={Mobgs: Motion deblurring dynamic 3d gaussian splatting for blurry monocular video},
  author={Bui, Minh-Quan Viet and Park, Jongmin and Gonzalez, Juan Luis and Moon, Jaeho and Oh, Jihyong and Kim, Munchurl},
  booktitle={Proceedings of the AAAI Conference on Artificial Intelligence},
  volume={40},
  number={4},
  pages={2480--2489},
  year={2026}
}

@inproceedings{keselman2022approximate,
  title={Approximate differentiable rendering with algebraic surfaces},
  author={Keselman, Leonid and Hebert, Martial},
  booktitle={European Conference on Computer Vision},
  pages={596--614},
  year={2022},
  organization={Springer}
}

@article{keselman2023flexible,
  title={Flexible techniques for differentiable rendering with 3d gaussians},
  author={Keselman, Leonid and Hebert, Martial},
  journal={arXiv preprint arXiv:2308.14737},
  year={2023}
}

@article{yu2024gaussian,
  title={Gaussian opacity fields: Efficient adaptive surface reconstruction in unbounded scenes},
  author={Yu, Zehao and Sattler, Torsten and Geiger, Andreas},
  journal={ACM Transactions on Graphics (ToG)},
  volume={43},
  number={6},
  pages={1--13},
  year={2024},
  publisher={ACM New York, NY, USA}
}

@inproceedings{yu2024mip,
  title={Mip-splatting: Alias-free 3d gaussian splatting},
  author={Yu, Zehao and Chen, Anpei and Huang, Binbin and Sattler, Torsten and Geiger, Andreas},
  booktitle={Proceedings of the IEEE/CVF conference on computer vision and pattern recognition},
  pages={19447--19456},
  year={2024}
}

@inproceedings{lin2024gaussian,
  title={Gaussian-flow: 4d reconstruction with dynamic 3d gaussian particle},
  author={Lin, Youtian and Dai, Zuozhuo and Zhu, Siyu and Yao, Yao},
  booktitle={Proceedings of the IEEE/CVF Conference on Computer Vision and Pattern Recognition},
  pages={21136--21145},
  year={2024}
}

@article{lin2025depth,
  title={Depth anything 3: Recovering the visual space from any views},
  author={Lin, Haotong and Chen, Sili and Liew, Junhao and Chen, Donny Y and Li, Zhenyu and Shi, Guang and Feng, Jiashi and Kang, Bingyi},
  journal={arXiv preprint arXiv:2511.10647},
  year={2025}
}

@article{yang2024depth,
  title={Depth anything v2},
  author={Yang, Lihe and Kang, Bingyi and Huang, Zilong and Zhao, Zhen and Xu, Xiaogang and Feng, Jiashi and Zhao, Hengshuang},
  journal={Advances in Neural Information Processing Systems},
  volume={37},
  pages={21875--21911},
  year={2024}
}

@inproceedings{xu2022gmflow,
  title={Gmflow: Learning optical flow via global matching},
  author={Xu, Haofei and Zhang, Jing and Cai, Jianfei and Rezatofighi, Hamid and Tao, Dacheng},
  booktitle={Proceedings of the IEEE/CVF conference on computer vision and pattern recognition},
  pages={8121--8130},
  year={2022}
}
}

\clearpage
\appendix
\section*{Appendix}

\section{Demo Videos}
\label{sec:demo}
We provide a demo video, \texttt{WebSpline\_demo.mp4}, with extensive qualitative comparisons between WebSpline and state-of-the-art novel view synthesis methods~\cite{splinegs, mosca, se3bsplinegs, worldtree}. The video shows results for novel view synthesis on the iPhone~\cite{gao2022monocular} and NVIDIA~\cite{yoon2020dynamic} datasets, as well as Gaussian trajectory visualizations on the iPhone dataset.

\section{Implementation Details}
\label{sec:implementation}
Our framework is built on the widely adopted open-source 3D Gaussian Splatting (3DGS) codebase~\cite{3dgs}. We obtain 2D point tracks using CoTracker3~\cite{karaev2025cotracker3}. For depth, we use the pre-trained UniDepth model~\cite{piccinelli2024unidepth} on the NVIDIA dataset~\cite{yoon2020dynamic} and sensor depth on the iPhone dataset~\cite{gao2022monocular}. WebSpline is trained for 2K and 20K iterations in the SPGF and SISO stages, respectively. 
In the SPGF stage, we linearly decay the current window length $|\mathcal{W}|$ from $N_f$ to $0.1 \times N_f$ to speed up optimization. The starting frame $t_1$ is sampled from $[0, N_f - |\mathcal{W}|]$ via a Beta distribution to provide balanced supervision across the entire sequence. We set $K=16$ in Eq.~\ref{eq:neighbor_and_edges} and $K_\text{sp}=8$ in Eq.~\ref{eq:spatial_neighbor}. We set the number of control points to $N_c=\lfloor N_f/4 \rfloor$ for the iPhone dataset~\cite{gao2022monocular} and $N_c=N_f$ for the NVIDIA dataset~\cite{yoon2020dynamic}. We set the temporal offset $\Delta$ in Eq.~\ref{eq:otr_loss} and the index offset $\tau$ between control points in Eq.~\ref{eq:str_loss} to 1. We set $\alpha$ and $\beta$ values in Eq.~\ref{eq:str_loss} to 0.1. For $\mathcal{L}_\text{SPGF}$ and $\mathcal{L}_\text{SISO}$, we set all $\lambda$ values in Eqs.~\ref{eq:loss_spgf} and~\ref{eq:loss_siso} to 1, respectively.

\section{Additional Analysis}
\label{sec:additional_analysis}

\subsection{Analysis on $\mathcal{L}_\text{OTR}$}
\label{subsec:otr_analysis}

The orientation-similarity-based temporal rigidity loss $\mathcal{L}_\text{OTR}$ is related to the ARAP-style rigidity regularization used in the scaffold-based methods~\cite{mosca, se3bsplinegs, worldtree}.
Both losses encourage neighboring proxy nodes to preserve local geometric relationships over time.
Unlike the original ARAP formulation, which treats neighboring relations uniformly, $\mathcal{L}_\text{OTR}$ uses the orientation similarity weight $w^{(i,j)}$ to adaptively emphasize neighboring node pairs that exhibit similar local rigid motion.
This weighting reduces the influence of neighboring nodes that are spatially close but follow different motion patterns.
In WebSpline, $\mathcal{L}_\text{OTR}$ is used to optimize the SPG as structural guidance for subsequent SIS optimization, rather than to directly drive Gaussian deformation or rendering.

\begin{wraptable}{r}{0.43\linewidth}
\vspace{-10pt}
\centering
\scriptsize
\caption{Effect of orientation-similarity weighting in $\mathcal{L}_\text{OTR}$.}
\label{tab:otr_analysis}
\vspace{-4pt}
\resizebox{\linewidth}{!}{
\begin{tabular}{lccc}
\toprule
\multicolumn{1}{c}{Method} & mPSNR$\uparrow$ & mSSIM$\uparrow$ & mLPIPS$\downarrow$ \\
\midrule
w/o OTR & 19.79 & 0.728 & 0.232 \\
OTR w/o $w^{(i,j)}$ & 20.00 & 0.736 & 0.225 \\
Ours & \first{20.11} & \first{0.738} & \first{0.222} \\
\bottomrule
\end{tabular}
}
\vspace{-8pt}
\end{wraptable}

As shown in Table~\ref{tab:otr_analysis}, using $\mathcal{L}_\text{OTR}$ without the orientation similarity weight $w^{(i,j)}$ already improves over removing $\mathcal{L}_\text{OTR}$, confirming the benefit of temporal rigidity for SPG refinement.
Adding $w^{(i,j)}$ further improves mPSNR from 20.00 to 20.11 and reduces mLPIPS from 0.225 to 0.222, indicating that adaptive weighting provides more reliable structural guidance. Overall, $\mathcal{L}_\text{OTR}$ serves as an effective refinement objective for the SPG, while the main contribution of WebSpline lies in the SPG-to-SIS framework that uses this refined structural foundation to learn structurally coherent yet flexible spline trajectories for dynamic Gaussians.

\subsection{Analysis on $\tau$}
\label{subsec:tau_analysis}

\begin{wraptable}{r}{0.43\linewidth}
\vspace{-10pt}
\centering
\scriptsize
\caption{Effect of the index offset $\tau$ between control points in $\mathcal{L}_\text{STR}$.}
\label{tab:tau_analysis}
\vspace{-4pt}
\resizebox{\linewidth}{!}{
\begin{tabular}{lccc}
\toprule
\multicolumn{1}{c}{Method} & mPSNR$\uparrow$ & mSSIM$\uparrow$ & mLPIPS$\downarrow$ \\
\midrule
$\tau=10$ & 19.97 & 0.736 & 0.225 \\
$\tau=7$  & 19.98 & 0.736 & 0.225 \\
$\tau=4$  & 20.02 & 0.737 & 0.224 \\
$\tau=2$  & 20.06 & 0.738 & 0.222 \\
Ours ($\tau=1$) & \first{20.11} & \first{0.738} & \first{0.222} \\
\bottomrule
\end{tabular}
}
\vspace{-8pt}
\end{wraptable}

We conduct an ablation study on the index offset $\tau$ between control points in $\mathcal{L}_\text{STR}$. As shown in Table~\ref{tab:tau_analysis}, the performance gradually decreases as $\tau$ increases. A larger $\tau$ enforces rigidity over a longer temporal span of control points, introducing long-range temporal constraints that can reduce the flexibility of SIS in modeling fine-grained Gaussian motion. Therefore, we set $\tau=1$ as the default setting.

\section{Additional Method Details}

This section provides additional details on the auxiliary components used in the SPGF stage. These components stabilize the SPG $\mathcal{K}$ before learning the SIS representation.

\subsection{Auxiliary Orientation Initialization}
\label{subsec:orient_init}
To initialize the auxiliary orientation sequence $\mathcal{O}^{(i)} = \{\mathbf{O}_t^{(i)}\}_{t=1}^{N_f}$, we employ a local Procrustes alignment strategy consistent with recent scaffold-based methods~\cite{mosca, se3bsplinegs, worldtree}.
Starting with $\mathbf{O}_1^{(i)} = \mathbf{I}$, we estimate the relative orientation update $\Delta \mathbf{O}_t^{(i)}$ for $t > 1$ by aligning the centered local configurations of neighboring nodes $j \in \mathcal{N}(i)$:
\begin{equation}
\Delta \mathbf{O}_t^{(i)} = \arg\min_{\mathbf{R} \in SO(3)} \sum_{j \in \mathcal{N}(i)} \left\| \mathbf{R} \tilde{\mathbf{a}}_{t-1}^{(i,j)} - \tilde{\mathbf{b}}_t^{(i,j)} \right\|_2^2,
\end{equation}
where $\tilde{\mathbf{a}}_{t-1}^{(i,j)}$ and $\tilde{\mathbf{b}}_t^{(i,j)}$ denote the mean-centered relative offsets of neighboring nodes at frames $t-1$ and $t$, respectively. We solve this using the standard SVD-based closed-form solution and accumulate the orientation as:
\begin{equation}
\mathbf{O}_t^{(i)} = \Delta \mathbf{O}_t^{(i)} \mathbf{O}_{t-1}^{(i)}.
\end{equation}
These orientations are exclusively used to compute the orientation similarity weights $w^{(i,j)}$ and local-coordinate offsets within $\mathcal{L}_\text{OTR}$. 

\subsection{Velocity and Acceleration Regularization}
\label{subsec:vel_acc_reg}

We apply standard temporal smoothness regularizations to the SPG node positions and auxiliary orientations~\cite{mosca, se3bsplinegs, worldtree}.
For each structural proxy node $v^{(i)}$, we regularize both the position sequence $\{\mathbf{q}^{(i)}_t\}_{t=1}^{N_f}$ and the auxiliary orientation sequence $\{\mathbf{O}^{(i)}_t\}_{t=1}^{N_f}$.

For the position sequence, the velocity and acceleration terms are defined as
\begin{equation}
\mathcal{L}^{\text{pos}}_\text{vel}
=
\sum_{t=1}^{N_f-1}
\sum_{i=1}^{M}
\left\|
\mathbf{q}^{(i)}_{t+1}
-
\mathbf{q}^{(i)}_{t}
\right\|_2,
\qquad
\mathcal{L}^{\text{pos}}_\text{acc}
=
\sum_{t=1}^{N_f-2}
\sum_{i=1}^{M}
\left\|
\mathbf{q}^{(i)}_{t+2}
-
2\mathbf{q}^{(i)}_{t+1}
+
\mathbf{q}^{(i)}_{t}
\right\|_2.
\end{equation}
For the auxiliary orientation sequence, we first compute the relative orientation update and its rotation angle between consecutive frames:
\begin{equation}
\Delta\mathbf{O}^{(i)}_t
=
\mathbf{O}^{(i)}_{t+1}
\left(\mathbf{O}^{(i)}_{t}\right)^\top,
\qquad
\theta^{(i)}_t
=
\angle\left(\Delta\mathbf{O}^{(i)}_t\right),
\qquad
t=1,\ldots,N_f-1,
\end{equation}
where $\angle(\cdot)$ denotes the rotation angle of an $SO(3)$ matrix.
The orientation velocity and acceleration terms are then defined as
\begin{equation}
\mathcal{L}^{\text{ori}}_\text{vel}
=
\sum_{t=1}^{N_f-1}
\sum_{i=1}^{M}
\theta^{(i)}_t,
\qquad
\mathcal{L}^{\text{ori}}_\text{acc}
=
\sum_{t=1}^{N_f-2}
\sum_{i=1}^{M}
\left|
\theta^{(i)}_{t+1}
-
\theta^{(i)}_{t}
\right|.
\end{equation}
Finally, the velocity and acceleration regularizations are given by
\begin{equation}
\mathcal{L}_\text{vel}
=
\mathcal{L}^{\text{pos}}_\text{vel}
+
\mathcal{L}^{\text{ori}}_\text{vel},
\qquad
\mathcal{L}_\text{acc}
=
\mathcal{L}^{\text{pos}}_\text{acc}
+
\mathcal{L}^{\text{ori}}_\text{acc}.
\end{equation}
These regularizations encourage smooth temporal evolution of the SPG in the SPGF stage, complementing $\mathcal{L}_\text{OTR}$.

\section{Additional Qualitative Results}
\label{sec:additional_visual_results}
\begin{figure}
    \centering
    \includegraphics[width=0.75\linewidth,keepaspectratio]{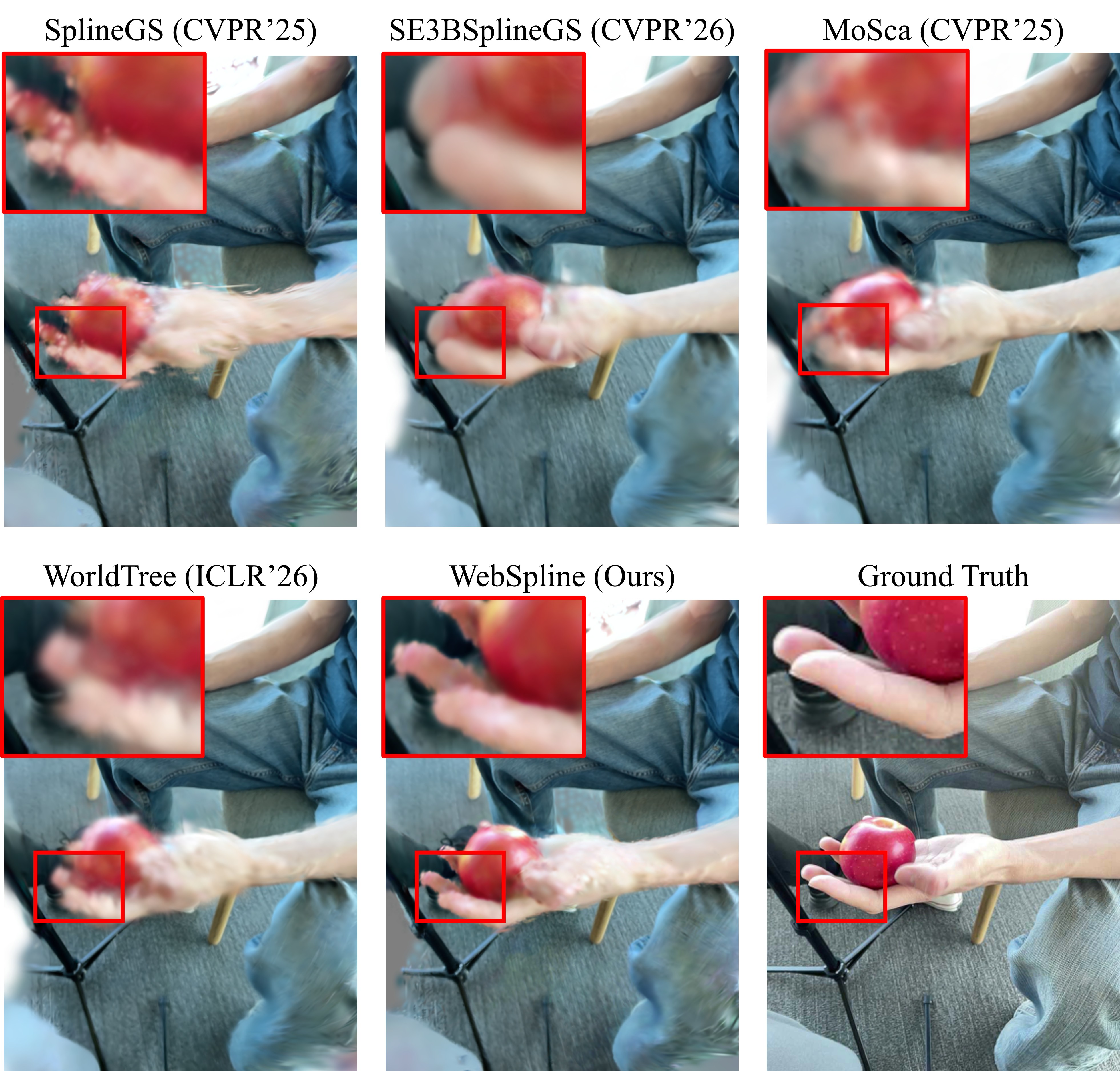}
    \caption{\textbf{Visual comparisons for novel view synthesis on the \textit{apple} scene from the iPhone dataset~\cite{gao2022monocular}.}}
    \label{fig:apple}
\end{figure}

\begin{figure}
    \centering
    \includegraphics[width=0.75\linewidth,keepaspectratio]{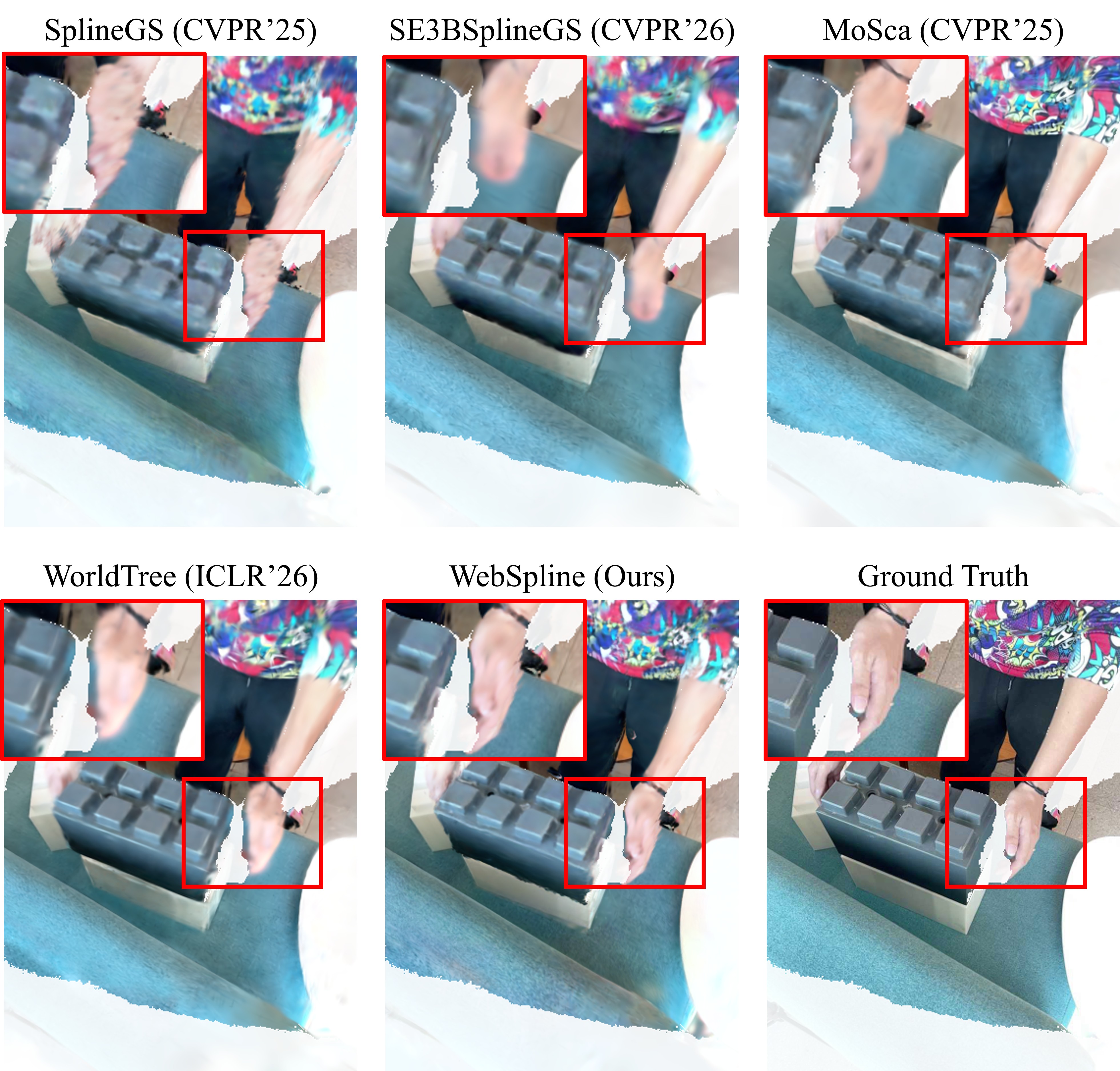}
    \caption{\textbf{Visual comparisons for novel view synthesis on the \textit{block} scene from the iPhone dataset~\cite{gao2022monocular}.}}
    \label{fig:block}
\end{figure}

\begin{figure}
    \centering
    \includegraphics[width=0.75\linewidth,keepaspectratio]{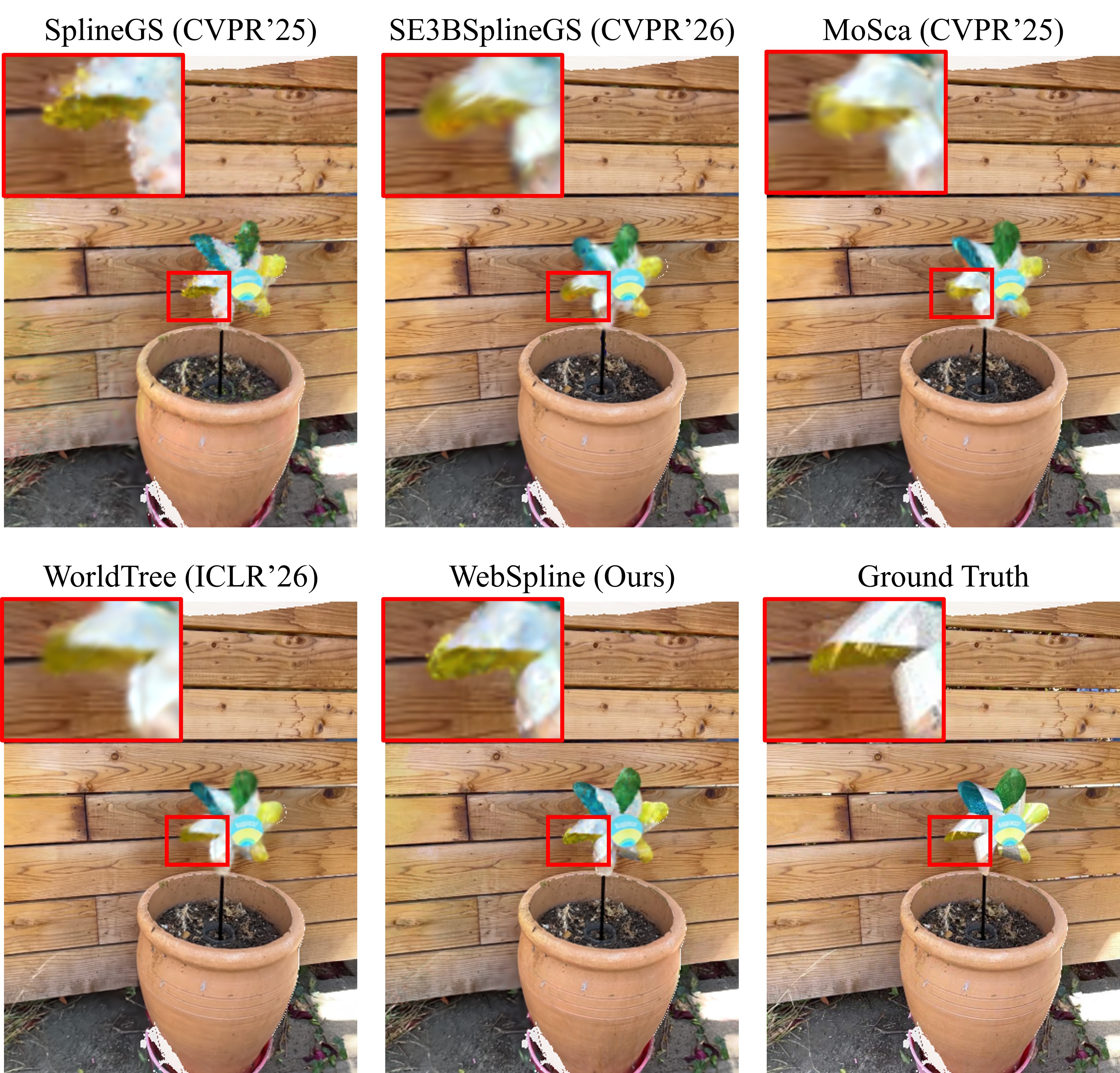}
    \caption{\textbf{Visual comparisons for novel view synthesis on the \textit{paper-windmill} scene from the iPhone dataset~\cite{gao2022monocular}.}}
    \label{fig:paper-windmill}
\end{figure}

\begin{figure}
    \centering
    \includegraphics[width=0.75\linewidth,keepaspectratio]{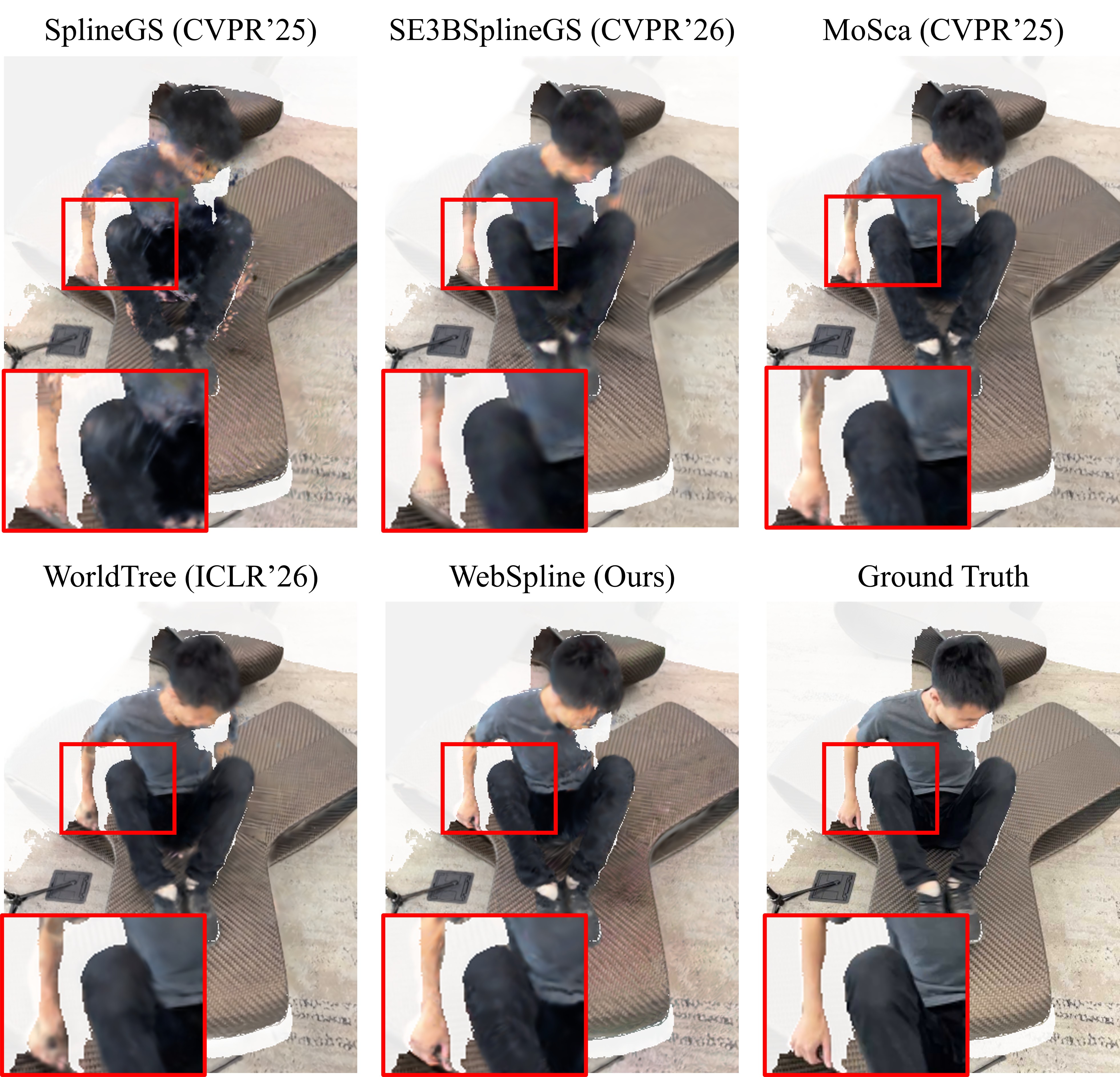}
    \caption{\textbf{Visual comparisons for novel view synthesis on the \textit{spin} scene from the iPhone dataset~\cite{gao2022monocular}.}}
    \label{fig:spin}
\end{figure}

\begin{figure}
    \centering
    \includegraphics[width=0.75\linewidth,keepaspectratio]{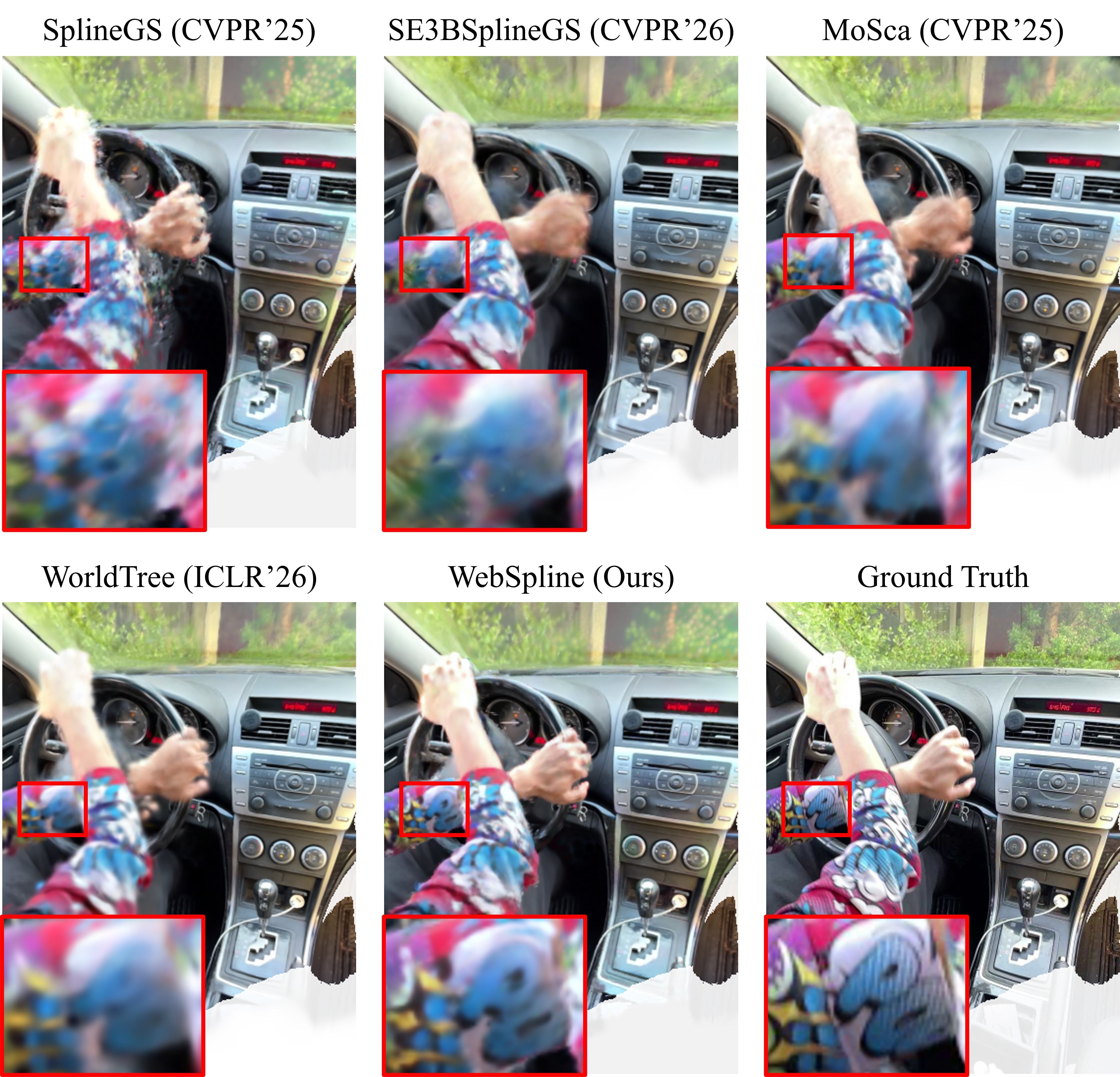}
    \caption{\textbf{Visual comparisons for novel view synthesis on the \textit{wheel} scene from the iPhone dataset~\cite{gao2022monocular}.}}
    \label{fig:wheel}
\end{figure}

Figs.~\ref{fig:apple}, \ref{fig:block}, \ref{fig:paper-windmill}, \ref{fig:spin}, and \ref{fig:wheel} present additional qualitative comparisons for novel view synthesis on the iPhone dataset~\cite{gao2022monocular}.

\section{Limitations and Future Work}
\label{sec:limitation}
Although WebSpline achieves high-quality dynamic scene reconstruction with fast rendering, it requires scene-specific optimization for each video sequence. This limits its applicability to immediate reconstruction scenarios. Extending WebSpline to a feed-forward framework that directly predicts the SIS representation for a given video sequence is an important direction for future work.

\section{Broader Impact}
WebSpline may support applications in AR/VR, telepresence, robotics, and digital content creation by enabling efficient dynamic 3D reconstruction from monocular videos. It can help reduce the need for specialized multi-camera capture systems and make dynamic 3D content creation more accessible. However, potential risks include privacy concerns when capturing people or private spaces, as well as the misuse of synthesized novel views. As such accessible rendering technologies advance, it becomes increasingly critical to establish robust consent protocols and clear disclosure mechanisms for synthesized visual content.

\clearpage

\end{document}